\documentclass[10pt,journal,compsoc]{IEEEtran}
\ifCLASSOPTIONcompsoc
  \usepackage[nocompress]{cite}
  
  \usepackage{caption} 
  \usepackage{subcaption}
  \usepackage{todonotes}
    
  \usepackage[ruled,vlined]{algorithm2e} 
  \usepackage{amssymb}
  \usepackage{mathtools}
    
  \usepackage{soul}
    
  \usepackage{color}

\else
  \usepackage{cite}
\fi
\ifCLASSINFOpdf
\else
\fi
\begin{document}
%
\title{Fitness Landscape Footprint: A Framework to Compare Neural Architecture Search Problems}
%
%
%
%

\author{Kalifou~Ren\'e~Traor\'e,
        \IEEEcompsocitemizethanks{\IEEEcompsocthanksitem Ren\'e Traor\'e is with the chair of Data Science in Earth Observation at the Technical University of Munich, 80333 Munich, Germany,
        and the Remote Sensing Technology Institute (IMF) of the German Aerospace Center (DLR), 82234 Wessling, Germany 
        (email:kalifou.traore@dlr.de).}
        Andr\'es~Camero
        \IEEEcompsocitemizethanks{\IEEEcompsocthanksitem Andr\'es~Camero is with Remote Sensing Technology Institute (IMF) of the German Aerospace Center (DLR), 82234 Wessling, Germany 
        (email:andres.camerounzueta@dlr.de).}
        and~Xiao~Xiang~Zhu,~\IEEEmembership{Fellow,~IEEE}
        \IEEEcompsocitemizethanks{\IEEEcompsocthanksitem ~Xiao~Xiang~Zhu is with the chair of Data Science in Earth Observation at the Technical University of Munich, 80333 Munich, Germany,
        and the Remote Sensing Technology Institute (IMF) of the German Aerospace Center (DLR), 82234 Wessling, Germany 
        (email:xiaoxiang.zhu@dlr.de).}
}

\markboth{Journal of \LaTeX\ Class Files,~Vol.~14, No.~8, August~2015}%
{Shell \MakeLowercase{\textit{et al.}}: Bare Demo of IEEEtran.cls for Computer Society Journals}
%



\IEEEtitleabstractindextext{%
\begin{abstract}
\emph{Neural architecture search} is a promising area of research dedicated to automating the design of neural network models.
This field is rapidly growing, with a surge of methodologies ranging from \textit{Bayesian optimization}, \textit{neuroevoltion}, to \textit{differentiable search}, and applications in various contexts. 
However, despite all great advances, few studies have presented insights on the difficulty of the problem itself, thus the success (or fail) of these methodologies remains unexplained.
In this sense, the field of \emph{optimization} has developed methods that \emph{highlight} key aspects to describe optimization problems.
The \emph{fitness landscape analysis} stands out when it comes to characterize reliably and quantitatively search algorithms.
In this paper, we propose to use \emph{fitness landscape analysis} to study a \emph{neural architecture search} problem. Particularly, we introduce the \emph{fitness landscape footprint}, an aggregation of eight (8) general-purpose metrics to synthesize the landscape of an architecture search problem. 
We studied two problems, the classical image classification benchmark CIFAR-10, and the Remote-Sensing problem So2Sat LCZ42.
The results present a quantitative appraisal of the problems, allowing to characterize the relative difficulty and other characteristics, such as the \emph{ruggedness} or the \emph{persistence}, that helps to tailor a search strategy to the problem. Also, the \emph{footprint} is a tool that enables the comparison of multiple problems. 
\end{abstract}

\begin{IEEEkeywords}
Neural Architecture Search, Fitness Landscape Analysis.
\end{IEEEkeywords}}

\maketitle

\IEEEdisplaynontitleabstractindextext

%
\IEEEpeerreviewmaketitle

\IEEEraisesectionheading{\section{Introduction}\label{sec:introduction}}

%
%
%
%
\IEEEPARstart{N}{eural architecture search} (NAS) has seen lots of advances in recent years~\cite{ojha2017review,elsken2019neural,ren2020comprehensive}.
State-of-the-art techniques 
have brought attention to~\textit{evolutionary algorithms}~\cite{EvolAging}
as well as to~\textit{differentiable search} strategies~\cite{liu2018darts,dong2019gdas} making NAS increasingly faster. Moreover, lots of efforts have been made to improve the reproducibility of research in the field,
by proposing benchmarks~\cite{ying2019nasbench101,dong2020nasbench201,siems2020nasbench301} 
and guidelines~\cite{Lindauer2020bestPractices, traore2021dso}.

However, despite all improvements made so far, 
limited insights have been presented explaining the success of state-of-the-art strategies or the difficulty of performing the search itself on a given task~\cite{Yang2020NasEvalHard}.
Surprisingly given enough budget it has been proven that simple baselines such as \emph{local search} are still top performers~\cite{white2019localsearch}.
This holds for various search spaces and datasets in \emph{computer vision} (CV)~\cite{Lindauer2020bestPractices}.

On the other hand, the \emph{optimization} field has explored and provided tools to address 
the issue of characterizing optimization problems. 
Among these, the \emph{fitness landscape analysis} (FLA)~\cite{watson2010introduction,pitzer2012comprehensive}, 
gives an intuitive idea of where (and how) the search is done
and can be improved. 
It has been used to study problems and to benchmark
algorithms in applications ranging from the field of physics, biology, 
to chemistry or pure mathematical optimization problems~\cite{combinatorial_fla_reidys}.  

\begin{figure}
    \centering
    \includegraphics[width=\columnwidth]{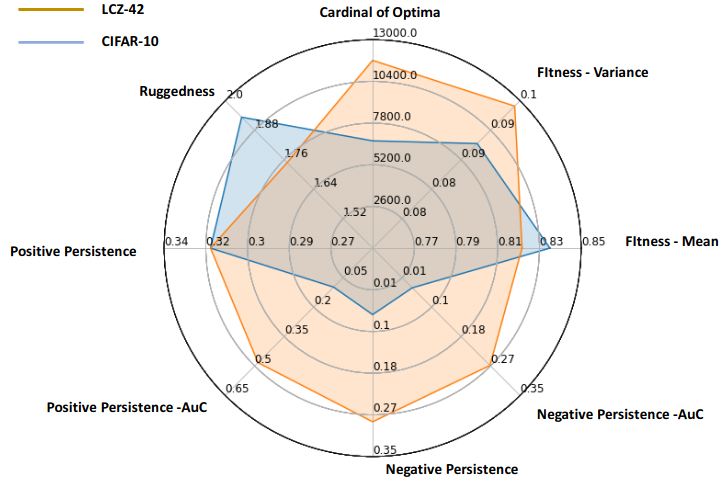}
    \caption{\emph{Fitness landscape footprint} of samples trained for 36 epochs on CIFAR-10 and So2Sat LCZ42.}
    \label{fig:footprints}
\end{figure}

 This study proposes to fill the gap by using FLA to quantitatively characterize 
 the difficulty of performing NAS on a given search space and setting. 
 Particularly, we propose to characterize the landscape of a NAS problem ($\digamma$) by relying on aspects such as the \emph{density of fitness}, \emph{fitness distance correlation (FDC)}, \emph{ruggedness}, and \emph{persistence}.
 Therefore, the main contributions of this study are:
 
 \begin{itemize}
    \item  We introduce the \emph{fitness landscape footprint}, 
    a framework to characterize the landscape of a NAS problem ($\digamma$). In practice, this framework helps compare 
    landscapes associated to various search spaces, fitness functions, neighborhood operators or even datasets.
    \item Using the \emph{fitness landscape footprint}, we studied two image classification datasets, the classical CIFAR-10~\cite{cifar10} as well as the real-world remote sensing problem, So2Sat LCZ42~\cite{So2SatDataset}.
    Among the findings, we identify several clues indicating that NAS could be performed at shorter regimen (36 epochs), finding elite models early.
    Other findings show a common signature of fitness of the search space on both datasets and the visualization of landscape of both problems 
    for various training settings.
\end{itemize}

The remainder of this paper is organized as follows. The next section briefly introduces the related work. Section~\ref{sec:proposal} presents our proposal. 
Section~\ref{sec:setup} outlines the experimental setup. Section~\ref{sec:results} presents the results. Section~\ref{sec:conclusions} summarizes the conclusions and proposes future work.

\section{Related Work}\label{sec:relatedwork}

This section introduces the related work in NAS as well as in fitness landscape analysis.

\subsection{Neural Architecture Search (NAS)}

In the late 1980s, researchers started exploring the use of \emph{evolutionary computation} to design and train neural networks,
a.k.a., \textit{neuroevolution}~\cite{engel1988fnn-sa,montana1989fnn-ga,alba1993genetic,alba1993full,yao1993review}. 
Neuroevolution gained popularity in the 2000s thanks to the \emph{NeuroEvolution of Augmenting Topologies} (NEAT) method~\cite{stanley2002evolving}. 
NEAT is a \emph{genetic algorithm} (GA) that increasingly evolves complex neural network topologies (structures) and weights. 
More recently in \cite{garciarena2021exploitation}, authors take advantage of information obtained during deployment of neuroevolutionary NAS algorithm, 
to inform future runs of search. 
This information is to be stored in a Bayesian network-based metamodel and used in the case of search of \emph{generative adversarial networks} (GAN) architectures in CV. 
The metamodel is shown to successfully guide a strategy based on \emph{local search}, to find competitive models on the task of generating images from the MNIST image benchmark.

With the rise of \emph{deep learning}, 
there has been a resurgence of these \emph{old} methods in recent years to tackle 
the complex task of neural network design~\cite{ojha2017review}. 
For example, some authors proposed to used \emph{harmony search}~\cite{rosa2015cnnhs}, GA~\cite{zhining2015genetic}, and \emph{mixed integer parallel efficient global optimization} technique~\cite{van2019automatic} to design \emph{convolutional neural networks} (CNNs). 
Related to \emph{recurrent neural networks} (RNNs), Bayesian optimization~\cite{camero2020bomrs}, neuroevolution~\cite{ororbia2019investigating}, and hybrid approaches~\cite{camero2019optlowcost}, has been explored. 
Also, NEAT has been extended to fulfill the current needs of CNN~\cite{miikkulainen2019evolving} and RNN~\cite{rawal2016evolving}.

On the other hand, recent methods such as DARTS~\cite{liu2018darts} take advantage of 
\emph{continuous optimization} to make search over a large graph of overlapping configurations (\emph{Super-Net}) differentiable.
These recent improvement result in large speedups in terms of search 
time~\cite{elsken2019neural} but also in some case in a lack of robustness and interpretability~\cite{Yang2020NasEvalHard}.

In an attempt to make NAS more reproducible and accessible to the research community,
additional efforts has been directed in providing open-source benchmarks of 
neural network configurations and their evaluations.
These have been spanning several areas of applied \emph{machine learning} (ML) such 
as CV~\cite{ying2019nasbench101,dong2020nasbench201,zela2020nasbench1shot1,siems2020nasbench301}  
or even \emph{natural language processing}~\cite{Klyuchnikov2020NasBenchNLP}.
Given such available resources, 
the barrier for prototyping as well as better understanding NAS are progressively lowered.


\subsection{Fitness Landscape Analysis}

When it comes to the study of optimization problems, 
\emph{fitness landscape analysis} (FLA)~\cite{pitzer2012comprehensive}
has been long used to characterize optimization problems, with applications ranging from \emph{Physics}, \emph{Biology}, to \emph{Chemistry} or pure \emph{Mathematical optimization}~\cite{combinatorial_fla_reidys}. 
At the conceptual level, FLA leverages knowledge of underlying structures of the tackled problems, while using general-purpose features to characterize them.
One of the aims of FLA is to help predict performances and tuning algorithms, when solving an optimization problem.
It does so by carefully considering appropriate descriptions of a fitness function, a search space and transition operators 
\cite{Stadler02bFL,Hernando2012CardOptimaBenchmark,Matthew99estimatingCardBigSS} to model a problem.
The use of descriptive features of the extracted landscapes include analysis of \emph{density of states},  
\emph{enumeration of optima}~\cite{Hernando2012CardOptimaBenchmark, Matthew99estimatingCardBigSS} and \emph{neutral sets} or \emph{basins}.
Another example of a general-purpose FLA feature is the \emph{fitness distance correlation}~\cite{FDC_Metric_Jones}, 
first applied to study the performances of a genetic algorithm on a white-box optimization problem. 
It aims at describing the hardness of a problem, using limited knowledge of an algorithm to be used. 
Its authors discuss a limitation in describing unknown functions to optimize for, i.e., in the case of black-box settings. 

FLA has been successfully applied to predict performance (or performance enhancement).
The work of \cite{LONs_FLA_ILS_Daolio} investigates the use of \emph{local optima networks} (LON) to model fitness landscapes of combinatorial problems using a graph. 
It studies the relationships between LON features and the performance of meta-heuristics such as \emph{iterated local search} algorithms. 
Their results show that some LON features can be used for the prediction of the running time to success.
%
In ~\cite{FSSP_FLA_CLERGUE2018449}, a general-purpose landscape-aware \emph{hill-climbing iterated local search} strategy is introduced
to solve the cellular automata problem of the \emph{6-states firing squad synchronization problem} (FSSP). 
In particular, an acceptance criterion (\emph{neutral hill climbing} or \emph{netcrawler}) based on the neutrality of the landscape is proposed, to prevent the search algorithm from being stuck in search spaces with large plateaus.
%

The \emph{value constraint satisfaction problem} (VCSP) framework is introduced in \cite{VCSP_FLA_Kaznatcheev} to represent fitness landscapes. 
The work aims at finding classes of fitness landscapes enabling \emph{local search} strategies to be tractable, i.e., solvable in polynomial time.
This is done by identifying useful properties of VCs to be used, such as tree-structured graphs and boolean VCSP settings. 
\cite{NK_FLA_Ochoa} is among the first topological and statistical analyses providing a characterization of combinatorial landscapes.
Its authors use the concept of inherent networks, borrowed from statistical physics, to better represent landscapes graphically.  
They focus on the case of ~$NK$-landscapes and analyze its attributes such as basins of attractions, their size, and clustering coefficients. 

Also, FLA has been used to characterize \emph{multi-objective optimization} (MO) problems. 
For instance, in~\cite{FLA_EMO_liefooghe} it is proposed to predict the performance of \emph{evolutionary multi-objective} (EMO) search algorithms on blac-box functions. 
The authors highlight that the use of manually designed features to analyze EMO performances is a common practice, 
and propose additional general-purpose features (\emph{ruggedness} and \emph{multi-modality}) to tackle some black-box MO problems. 
%
In ~\cite{DMOPs_FLA_Essiet}, the focus is put on \emph{dynamical multi-objective optimization} problems (DMOPs) with a time-dependent MO fitness landscape. 
They propose a landscape-aware dynamical version of a classical algorithm (NSGA-III), 
using adaptive mutation and recombination operators in mating. 
The aim is to keep track of the moving MO fitness landscape, i.e., the Pareto front.

Recently, some authors have study ML problems using FLA.  
For example, \cite{Bosman2017FitnessLA_ErrorNNs} addresses the case of complexity reduction of neural networks using cost penalty regularization for weight elimination, by studying the change in the error landscape of the regularized neural networks.
Particularly, the study proposes hyperparameter changes to the regularization, to make the error landscape more maneuverable to the back-propagation algorithm used in training.
In~\cite{FLA_NeuroEvo_NN_Rodrigues}, they study the generalization ability of CNNs on various ML problems.
A strong emphasis is made to provide insight on tuning \emph{neuroevolution} algorithms to fit the tasks. 

When it comes to the particular topic of NAS, the exploitation of FLA is nascent.  
In ~\cite{FLA_AutoML_Pimenta}, authors propose to represent an \emph{AutoML} pipeline in the framework on FLA,
and analyze its heterogeneous search space using measures of FDC and neutrality ratio. 
\cite{FLA_NAS_GNNS_Nunes} is among the first works to use FLA features to characterize the search space explored by NAS methods for automatically designing \emph{graph neural networks} (GNNs). 
They use established metrics (\emph{fitness distance correlation}, \emph{dispersion}) to characterize the difficulty for NAS on several GNNs benchmarks. 

\section{Fitness Landscape Footprint}\label{sec:proposal}

This section introduces the~\emph{fitness landscape footprint}.
First, the fitness landscape~$\digamma$ is defined.
Then, we introduce general-purpose features for FLA, including the \emph{fitness distance correlation}, 
\emph{local optima} and the \emph{ruggedness}.
Next, we propose the concept of \emph{persistence} to characterize over time the behavior of sampled configurations.
Last but not least, we introduce the \emph{fitness footprint} framework
to describe and compare NAS optimization problems in a simple manner. 


\subsection{Fitness Landscape}

A \emph{fitness landscape}~$\digamma$ is a framework to help study any optimization problem defined by a \emph{search space~$\Omega$}, 
a \emph{measurement of fitness~$f$} for samples in~$\Omega$, and a \emph{neighborhood operator~$N$} to navigate~$\Omega$.
It is defined as the triplet combination~$\digamma=( \Omega, f, N)$~\cite{Stadler02bFL}.

In a landscape, the fitness function $f$ assigns a fitness value to every configuration $x$ in the search space $\Omega$:
\begin{equation}
\begin{multlined}
 f:\Omega \longrightarrow \mathrm{R} \\
 x \longmapsto f(x)\\
\label{eq-fitness}
\end{multlined}
\end{equation}{}
\vspace{-0.6cm}

\noindent In the context of NAS, 
we consider $f$ as being the evaluation of performance
for a neural network configuration $x \in \Omega$ 
after a training regime of length $t_i$.

In order to provide a structure to the fitness landscape, 
one needs to think of a way to arrange the configurations of the search space ~$\Omega$,
and how they can be reached from one another. 
This is the role of the neighborhood operator~$N$ which assigns to each solution~$x$ in the
search space~$\Omega$ a set of neighbors~$N(x)\in P(\Omega)$.
%
For our NAS use case, we consider the \textit{Hamming distance} as the metric to define a neighborhood~$N$ 
as shown in  Equation~\ref{eq-neiborhood-hamming}:
\begin{equation}
\begin{multlined}
 N(x) = \{y \in \Omega \mid d_{hamming} (x, y) =1 \} \\
\label{eq-neiborhood-hamming}
\end{multlined}
\end{equation}{}
\vspace{-0.5cm}

\noindent where for each configuration~$x \in  \Omega$, its neighbors~$N$ comprise all configurations~$y \in  \Omega$ 
distant of a Hamming unit (1) to~$x$.

\subsection{Fitness Distance Correlation (FDC)}
The \emph{fitness distance correlation}~\cite{FDC_Metric_Jones, pitzer2012comprehensive} is a classical FLA concept used to characterize the hardness of optimization problems.
Originally used as a general-purpose score to study fitness landscapes,
it can also be visually assessed as the fitness versus distance to a global optimum~$x^*$, for all solutions~$y \in \Omega$.  
Equation~\ref{eq:landscape-profile} introduces the concept: 
\begin{equation}
FDC(x^*,\Omega, f) =
\{ (d_{hamming} (x^*, y), f(y) ) \textrm{, } \forall y \in \Omega 
\}
\label{eq:landscape-profile}
\end{equation}
\noindent
where~$x^*$ is the global optimum,~$\Omega$ is the search space, $f$~is a fitness function, and~$d_{hamming}$ 
is the~\emph{hamming distance}.

\subsection{Ruggedness}\label{subsec:ruggedness}
In the context of NAS, a \emph{random walk} represents a path of consecutive random steps 
in the space of neural networks representations, where for each pair of consecutive solutions $d_{hamming} (x^i, x^{i+1}) =1$. 
For instance, when configurations are represented by a binary vector, 
a step of random walk consist in the action of randomly flipping a bit in the vector, resulting in a new state. 

In this context, an additional feature to characterize the difficulty of optimizing over a fitness landscape~$\digamma$ 
is to measure its~\emph{ruggedness}~\cite{Stadler02bFL}.
A common metric of ruggedness is the autocorrelation~$\rho$ (serial correlation) 
on a series of fitness values $f$ for configurations in a random walk~$\mathrm{W}=\{ x^{0},..., x^{j},...,x^{n} \}$ in~$\Omega$. 
Once~$\rho$ is derived, the final ruggedness $\tau$ is the inverse of the autocorrelation for all consecutive samples (~$k=1$ lags):
~$\tau = \frac{1}{\rho(1)}$.
Equation~\ref{eq:acf} introduces the formula of the autocorrelation function~$\rho$:
\begin{equation}
\begin{multlined}
\rho(k) = \frac{
\mathrm{E}[(f(x_i) -\Bar{f})(f(x_{i+k}) -\Bar{f})] 
}{ Var( f(x_i) )}
\textrm{,} \forall i \in \mid \mathrm{W} \mid\\
\end{multlined}
\label{eq:acf}
\end{equation}{}
\vspace{-0.5cm}

In NAS, measuring the~ruggedness of~$\digamma$ can be interpreted as the 
variability in fitness~$f$ one can expect from a local search baseline algorithm in~$\Omega$.

\subsection{Local Optima}

Combinatorial optimization problems often aim at finding 
solutions either maximizing or minimizing 
a cost function $f$, in the case of single-objective problems\cite{pitzer2012comprehensive}.
Therefore, looking for such optimal configurations 
should not be diverted by the existence of locally optimal solutions.

Considering a fitness function~$f$, and a neighborhood structure~$N$,
a configuration~$x^*$ in search space~$\Omega $ is a local optimum (minimum)
if its fitness is lower than any of its neighbors, i.e., ~$f(x^*) \leq f(y),$ 
$\forall y \in N(x^*)$. In this case, $x^*$ is a \emph{local minimum}. 
In the case of a fitness lower 
than any other solution in the search space, 
i.e., $f(x^*) \leq f(y)$
, $\forall y \in \Omega$, then~$x^*$ is the \emph{global minimum}. The same can be defined for \emph{local maximum} and \emph{global maximum}.

Moreover, a way to measure the difficulty of a fitness landscape~$\digamma$ is to enumerate the existing local optima in~$\Omega$.  
On the individual basis, a local optimum can be retrieved using a local search procedure~\cite{Hernando2012CardOptimaBenchmark}, 
e.g., a \emph{best-improvement local search}  (BILS) for a local maximum. 
The algorithm~\ref{algo:best-imp-local-search} describes the procedure of BILS.

\begin{algorithm}[ht]
\SetAlgoLined
Choose an initial solution $x \in \Omega$;\\
\Repeat{$x=x^*$}{
    $x^*=x$\;
    \For{$i=1 \textbf{   to   } \mid N(x^*) \mid$}{
        Choose $y_{i} \in N(x^*)$\;
        \If{ $f(y_{i}) < f(x)$}{
          $x=y_{i}$\;
        }
    }
}
\caption{Best-improvement local-search (BILS)}
\label{algo:best-imp-local-search}
\end{algorithm}

Several methodologies have been defined to estimate the number of optima in $\Omega$~\cite{Hernando2012CardOptimaBenchmark}.
One of the simplest, and least computationally expensive, is the \emph{birthday problem}~\cite{Matthew99estimatingCardBigSS}.
As described in Algorithm~\ref{algo:local-optima-enumaration},
the procedure consist in an average estimation over~$T$ trials.
For each trial~$i$, we collect~$M$ local optima by applying BILS from~$M$ distinct and randomly selected starting points. 
Then, we measure~$k_i$, the number of distinct local optima, until the first duplication out~$M$ samples at trial~$i$.
Then, we derive~$k_{mean}$ as the average rank of first duplicate for all~$T$ trials.

\begin{algorithm}[ht]
\small
\SetAlgoLined
Let $T$ the total number of trials of enumeration;\\
\For{$i=1 \textbf{   to   } T$}{
    Choose $M$ distinct random starting points in $\Omega$;\\
    Iteratively collect the $M$ Local Optima using BILS;\\
    Let $k_i$ the number of Optima at first duplication;\\
}
Let $k_{mean}$ the average number of Optima at duplication;\\
Derive $N$ the number of Optima using $k_{mean}$ and Eq.~\ref{eq:card_opt};

\caption{Analytics of the birthday problem}
\label{algo:local-optima-enumaration}
\end{algorithm}

Equation \ref{eq:card_opt} describes how to obtain the 
final estimation of number of optima ~$N$:
\begin{equation}
N \approx \frac{k_{mean}^2}{-2 * ln(1 - P_D)}
\label{eq:card_opt}
\end{equation}
where $P_D=0.5$ is the fixed probability of duplicated local optima, and 
$k_{mean}$ the average number of (different) local optima found until the first duplication.
In other words, if $k_{mean}$ configurations are observed on average,
with a chance $P_D=0.5$ that two or more of them share the same rank,
then the number of local optima approximates to $N$.
It is analogous to the original \emph{birthday problem}, where one tries to estimate
the number $k_{mean}$ of persons that can fit in a room until two or more share the same birthday 
(given a fixed chance of duplicates $P_D=0.5$ and $N=365$ days in a year).
In our case, N is unknown, so we are tackling the \emph{reverse birthday problem} (a.k.a., the \emph{estimation of the martian year length}).

\subsection{Persistence}\label{subsec:persistence}

NAS aims to find neural network configurations
maximizing fitness~$f$ on unseen data (test sets)
after a given budget of training time~$t_i$.
While performances are usually only considered at the end of the training time~$t_i$,
we propose to study configurations based on how their fitness evolve thought training time,
given a series of fitness measurement~$\{f(t_0), ..., f(t_i), ..., f(t_n)\}$ 
and respectively increasing training budget~$\{t_0, ..., t_i, ..., t_n\}$.

Let us consider a ranking function~$Ranking(N, t_i)$, that retrieves all models~$y \in \Omega$,
with respect to a rank~$N$, based on $f(y)$ at training time~$t_i$; 
and an ordered series of training duration $R_{all}=\{t_0, ..., t_i, ..., t_n\}$.

Then, we define \textit{persistence}~$\Pi$ as the probability for model configurations 
ranked by function~$Ranking(\cdot)$ to keep their initial rank (i.e., the one observed at $t_0$) through $R_{all}$.
Equation~\ref{eq:persistence} outlines $\Pi$:
\begin{equation}
\Pi(Ranking(\cdot), N)  = 
\frac{
P( \cap_{t_i \in \mathrm{R_{all}}} Ranking(N, t_{i}) 
}{ P(Ranking(N, t_{0})) }\\
\label{eq:persistence}
\end{equation}{}

Particularly, we consider $Ranking(\cdot)$ to be either $Top\-Rank(\cdot)$, 
i.e., retrieving all models ranked $TOP-N (percentiles)$, 
or~$Bottom\-Rank(\cdot)$, for the $Bottom-N (percentiles)$ performers at training time $t_i$.
Then, the positive persistence~$\Pi_{Positive}(N)$ and $\Pi_{Negative}(N)$ informs on the probability 
for configurations in $\Omega$ to remain~$Top-N$ or $Bottom-N$ performers over time, respectively.
%
\begin{equation}
\begin{multlined}
\Pi_{Positive}(N) = \Pi(Top\-Rank(\cdot),N)\\
\Pi_{Negative}(N) = \Pi(Bottom\-Rank(\cdot),N)\\
\end{multlined}
\label{eq:persistence-pos-neg}
\end{equation}{}
\vspace{-0.5cm}

We also propose to measure the \emph{area under the curve} of the \emph{persistence}.
This metric shall inform on the evolution of the persistence as a function of the rank~$N$ (Equation~\ref{eq:persistence-auc}).
\begin{equation}
AuC(\Pi(\cdot, N))  = \int_{1}^{N_{max}} \Pi(\cdot, k) \,dk
\label{eq:persistence-auc}
\end{equation}{}
\vspace{-0.5cm}

\subsection{Fitness Landscape Footprint}\label{subsec:fit-footprint}

In order to provide researchers with a tool that helps them to characterize, and analyze the potential and difficulty of a NAS problem, and to compare different problems (including different landscapes, fitness functions, neighborhood operators, and search spaces), 
we propose the \textbf{fitness landscape footprint}. 
The \emph{footprint} is defined as a set of the following key quantities, derived from the analysis of the landscape:
\begin{itemize}
    \item \textit{Overall fitness}: A measure of the \emph{expected overall fitness}~$f$, 
    and its \emph{standard deviation} for all sampled configurations~$y \in \Omega$.  
    These inform on \emph{how easy} it is to fit the ML task at hand with search space~$\Omega$.
    \item \textit{Ruggedness}: A measure of the difficulty of performing 
    \emph{local search} in~$\Omega$ via analysis of \emph{random walks} (Section~\ref{subsec:ruggedness}). 
    In the case of several available walks, we select the ruggedness~$\rho_{mean}(1)$ 
    as the average of~$\rho_{i}(1)$ for all evaluated routes~$i$ on a given dataset~$d$. 
    The final measurement of the ruggedness~$\tau_{d}$ is obtained as~$\tau_{d}=1 / \rho_{mean}(1)$.    
    A large ruggedness would imply large fluctuation (little correlation) from one step to the other of a local search. Little values indicate smoothness (high correlation) in fitness. 
    \item \textit{Cardinal of optima}: An estimation of the number of \emph{local optima} in~$\Omega$ (Algorithm~\ref{algo:local-optima-enumaration}). 
    \item \textit{Positive \& negative persistence}: The probabilities $\Pi$ for model configuration~$y \in \Omega$ 
    to remain among the best (or worse) over time (Section~\ref{subsec:persistence}).
    In particular, the probability~$\Pi(N=25)$ of keeping a Q1 rank($Top$ or~$Bottom$~$-25\%$), and
    the \emph{area under the curve} $AuC(N=25)$ of the persistence for~$ N =< 25$. 
    While the measurement of persistence at Q1 would inform on the chance of getting an elite model,
    its~$AuC$ (see Equation~\ref{eq:persistence-auc}) would tell us about the evolution 
    of such persistence at more restricted ranks ($N<25$).
\end{itemize}

\section{Experimental Setup}\label{sec:setup}

Our experiments aim to describe and compare  the landscape of NAS problems using the~\emph{fitness footprint}.
This section describes the datasets used to derive NAS landscapes, and provides 
additional details regarding the experiments. 

\subsection{Data Sets}

\noindent \textit{CIFAR-10}
is an image classification data set~\cite{cifar10}.
It consists of 60000 images (32x32 pixels) and its correspondent label (with no overlap). The data is split in 10 classes (6000 images per class), namely: airplanes, cars, birds, cats, deer, dogs, frogs, horses, ships, and trucks.
Also, the data is split into training (50000 samples) and test (10000 samples).

\begin{figure}[h]
    \centering
    \includegraphics[width=\columnwidth]{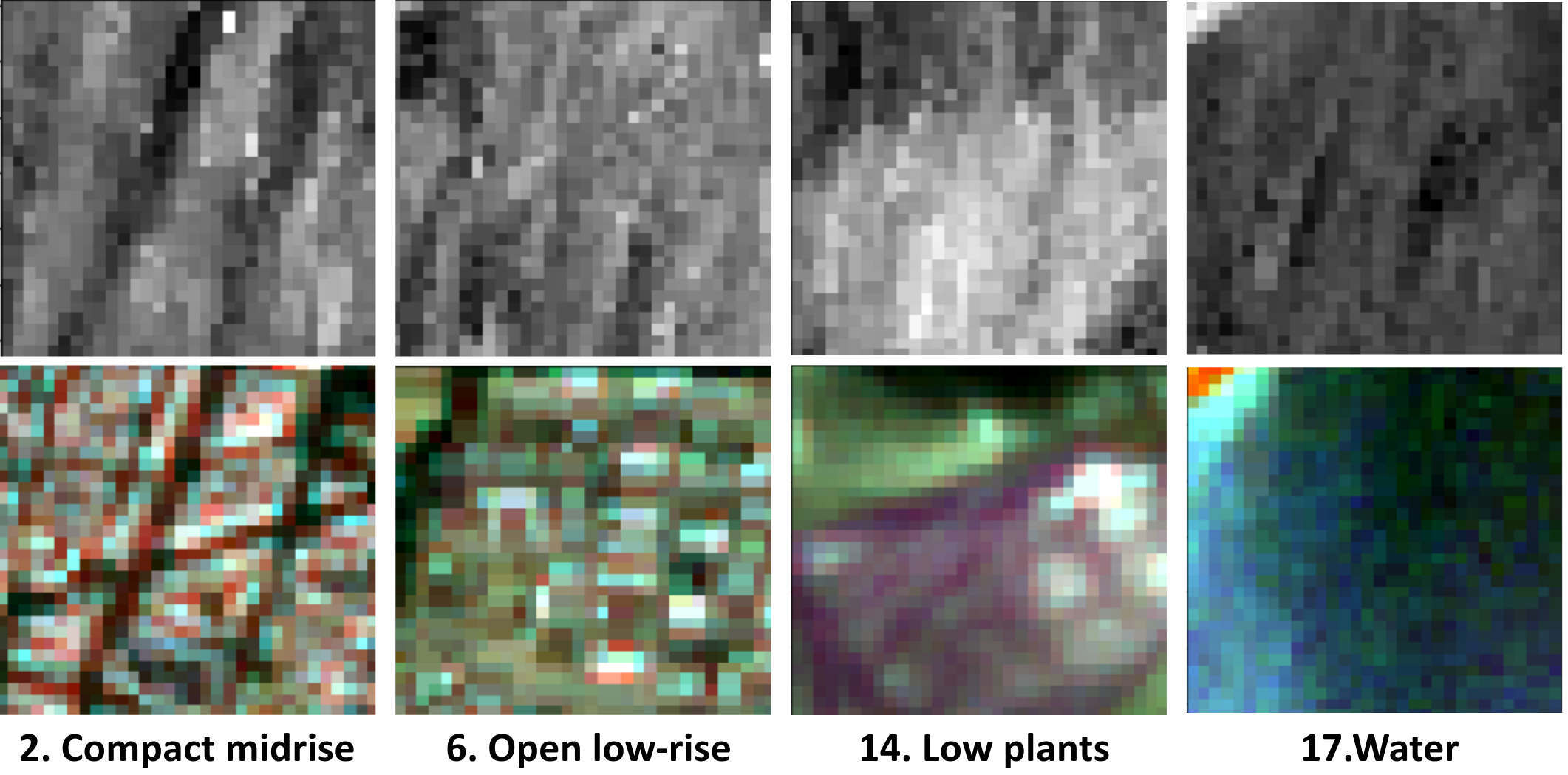}
    \caption{So2Sat LCZ42 samples.  The top row contains SAR patches (Sentinel-1), followed by the associated Multi-spectral patches (Sentinel-2) in the bottom row.}
    \label{fig:lcz42-dataset-visual}
\end{figure}

\noindent
\textit{The So2Sat LCZ42} is an \emph{Earth observation} image classification dataset~\cite{So2SatDataset}. 
It contains co-registered image patches from the Sentinel-1 and Sentinel-2 satellite sensors, 
all assigned to a single label out of 17 classes. 
Each pair is of 32x32 pixels, 10 multi-spectral bands for Sentinel-1 images and 8 bands for Sentinel-2. 
The existing classes describe each image patch in the global land cover framework of \emph{local climate zones} (LCZs), 
with 10 classes assigned to urban or built areas (1 to 10) and 7 to natural sites (11 to 17).
The classes are as follows: compact high-rise (1), compact mid-rise (2), compact low-rise (3), open high-rise (4), open mid-rise (5), open low-rise (6), lightweight low-rise (7), large low-rise (8), sparsely built (9), and heavy industry (10), dense trees (11), scattered tree (12), bush, scrub (13), low plants (14), bare rock or paved (15), bare soil or sand (16), and water (17). Note that in the original data set the classes 11 to 17 are related as A to G.
Figure~\ref{fig:lcz42-dataset-visual} displays four (4) pairs of Sentinel-1 and Sentinel-2 image patches respectively from classes 
2, 6, 14 and 17.  

The data is split into training (352366 images), validation (24188) and test (24119).
It is important to note that two various pools of cities were used to build So2Sat LCZ42: samples collected from 32 cities 
around the globe were selected to form the training set, while samples from 10 other cites were used for 
the validation and test set, with a geographical split (east and west) to insure distinct samples.

\begin{figure}[h]
    \centering
    \includegraphics[width=\columnwidth]{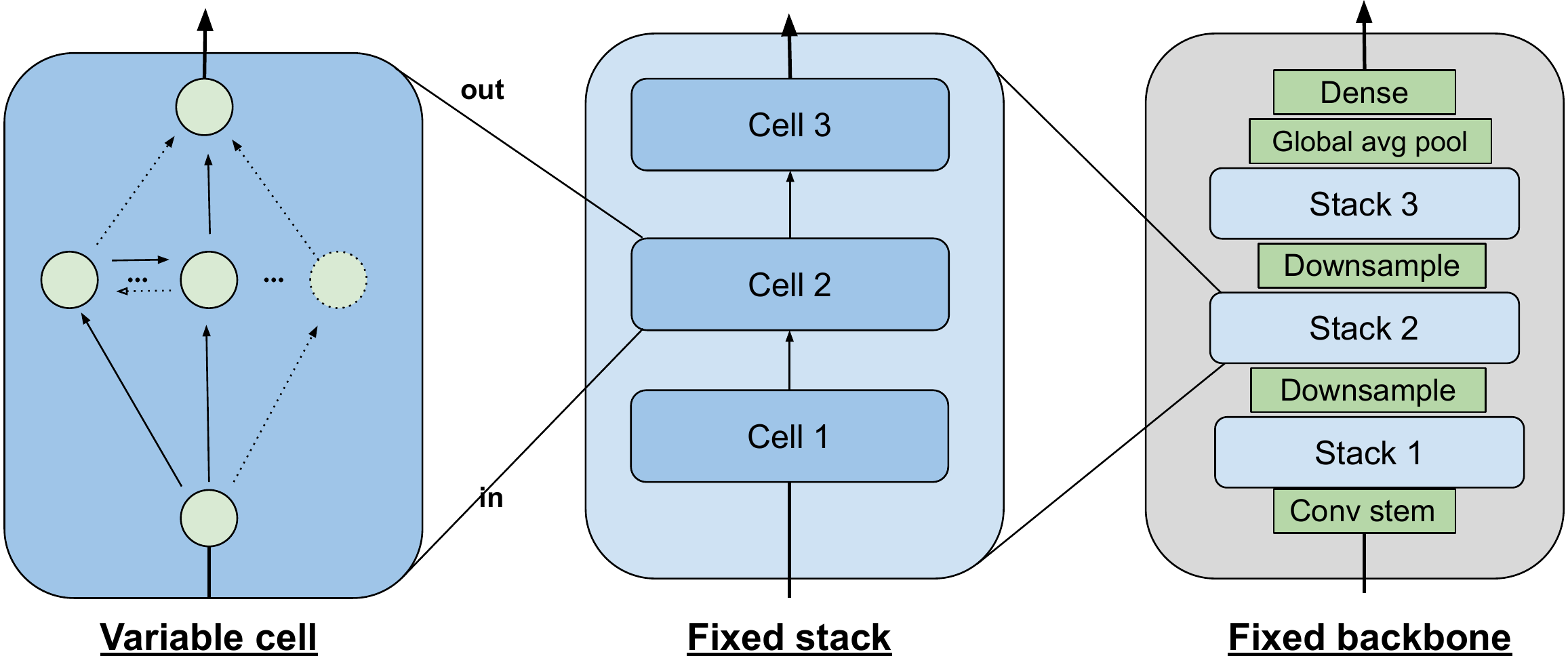}
    \caption{Visual description of the backbone of image classifier samples in NASBench-101.}
    \label{fig:nasbench-101-backbone}
\end{figure}

\noindent \textit{The NasBench-101}\label{subsec:setup-nb-101}
is a benchmark for NAS methodologies in the context of image classification~\cite{ying2019nasbench101}.
It consists of $N_{sol}=453k$ distinct CNN architectures and their fitness evaluation at various training steps (4, 12, 36 and 108 epochs), on the dataset of CIFAR-10.
Its underlying search space~$\Omega$ is of CNN configurations with a fixed backbone consisting of a head, a  sequence of three (3) repeated blocks, followed by a dense softmax decision layer.  
The head (conv stem) is a 3 x 3 convolution with 128 output channels.
Each block or stack, is a sequence repeating three (3) times an elementary unit, referred to as~\emph{cell}. 
In this~\emph{micro search-space}, a cell is a DAG containing at most $V=7$ nodes and $N_{edges}=9$ edges. 
Moreover, each node has a label selected out of $L=3$ possibilities: \{3x3 convolution, 1x1 convolution, 3x3 maxpool\}.
In practice, a solution~$x \in \Omega$ is represented by an adjacency matrix of variable size and its list of operators. 
Figure~\ref{fig:nasbench-101-backbone} illustrates the structure of the image classification backbone used in \textit{The NasBench-101}.

\subsection{Additional details}

The following paragraphs provide experimental details.
First, we explain the custom encoding used to represent samples involved in the FLA of both CIFAR-10 and So2Sat LCZ42.
Then we detail the sampling of the search space, the performance evaluation speed-up, data distribution and others, only involved in So2Sat LCZ42 experiments.

\emph{Custom encoding:} 
In order to analyze the landscape of both datasets, 
we use an alternative encoding to~\cite{ying2019nasbench101}, for solutions of the search space.
In the original DAG, there are at most five (5) intermediate nodes (excluding IN and OUT), each labelled using~$L$.
In the new DAG, we consider non-labelled nodes by replacing each with three (3), for each possible state of operator in~$L$.
The new graph has a fixed number of nodes, $V=1+5*3+1=17$ and the resulting adjacency matrix is non upper-triangular and of size $S=17*17=289$.
We decide to identify each candidate by the binary vector obtained after flattening the matrix.

\begin{figure}[h]
    \centering
    \includegraphics[width=.85\columnwidth]{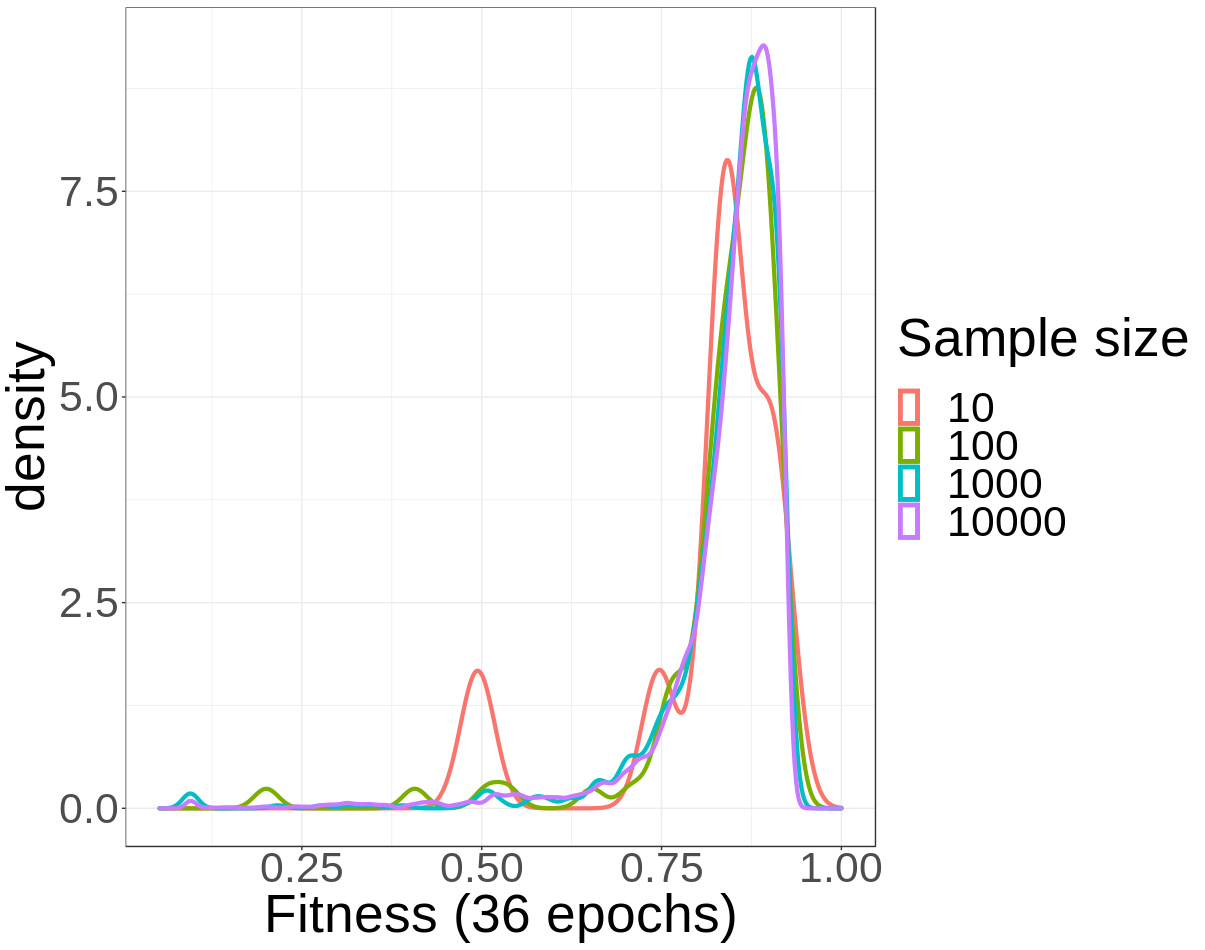}
    \caption{Density of fitness for various sizes of samples of CIFAR-10. }
    \label{fig:c10-val36-samp}
\end{figure}

\emph{Sampling of the search space:} 
Training numerous models can be expensive (time and 
hardware resources), thus
we identified the smallest number of sample to consider for representative results. 
Figure~\ref{fig:c10-val36-samp} displays the density of fitness (on validation, 36 epochs of training on CIFAR-10) for various sample sizes. The curves represent the fitness density for randomly selected models.
Based on this information, we set $N=100$ samples, because it presents a good trade-off between quality and number of evaluations.
In practice, we use a \emph{Latin hypercube sampling} (LHS) strategy to draw the samples (to have a well distributed sample).
Because of the complexity of sampling on the larger encoding, we perform LHS on the joint representation of original adjacency matrix and list of operators.

\emph{Data distribution:} 
Regarding experiments on So2Sat LCZ42, we split the original training set randomly into \emph{final training} (80\%) and \emph{final test} (20\%). 
The motivation is to ensure training and evaluation of models on the same data distribution, 
as done in CIFAR-10. For each data sample, we only consider Sentinel-2 as source of input.

\begin{figure}[h]
    \centering
    \includegraphics[width=0.9\columnwidth]{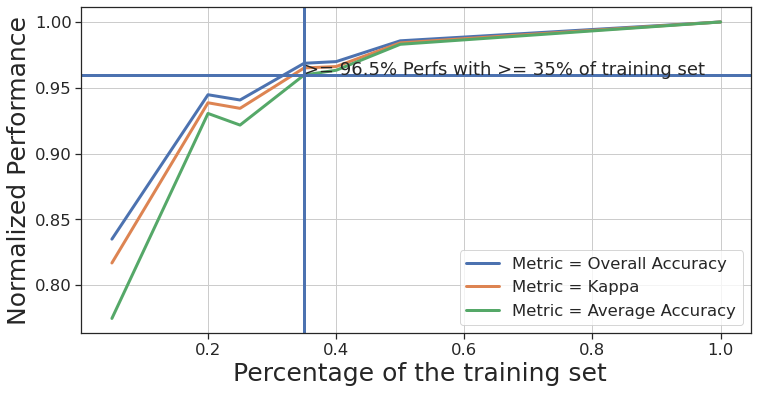}
    \caption{Fitness as a function of the share of training data on So2Sat LCZ42.    
    }
    \label{fig:lcz42-training-subset}
\end{figure}

\emph{Performance evaluation speed-up:} 
In order to speed up the evaluation of models on So2Sat LCZ42, 
we look for the smallest subset of training data enabling representative fitness in test. 
Figure \ref{fig:lcz42-training-subset} shows fitness in test as a function of the share (\%) of the training set, assessed with the \emph{overall accuracy}, \emph{kappa Cohen coefficient}, and \emph{average accuracy}. 
The values are normalized with respect to using 100\% of the training set.
We identify that using 35\% of the training set enables to reach 96.5\% of the reference performance. Thus, we ran our experiments using this setting.
For each model, we consider the fitness as the average fitness of three independent runs.

\emph{Miscellaneous:} %
Due to memory limitations, we set the size of the batches to 128 (instead of 256) 
when training or evaluating on So2Sat LCZ42.
The rest of the experimental hyper-parameters are default as in~\cite{ying2019nasbench101}. 
The training and inference were done on a Nvidia V100 GPU.


\section{Results}\label{sec:results}

This section introduces the experimental results. 
First, we study the fitness distribution for both problems.
Then, we analyze the landscape characteristics, such as the \emph{fitness distance correlation}, the \emph{ruggedness} and \emph{local optima}.
Next, we present the \emph{persistence} results. 
Finally, we \emph{assess and compare} the respective \emph{fitness footprints}.

\subsection{Density of Fitness}

\begin{figure}[h]
  \begin{subfigure}{\linewidth}
  \includegraphics[width=.45\linewidth]{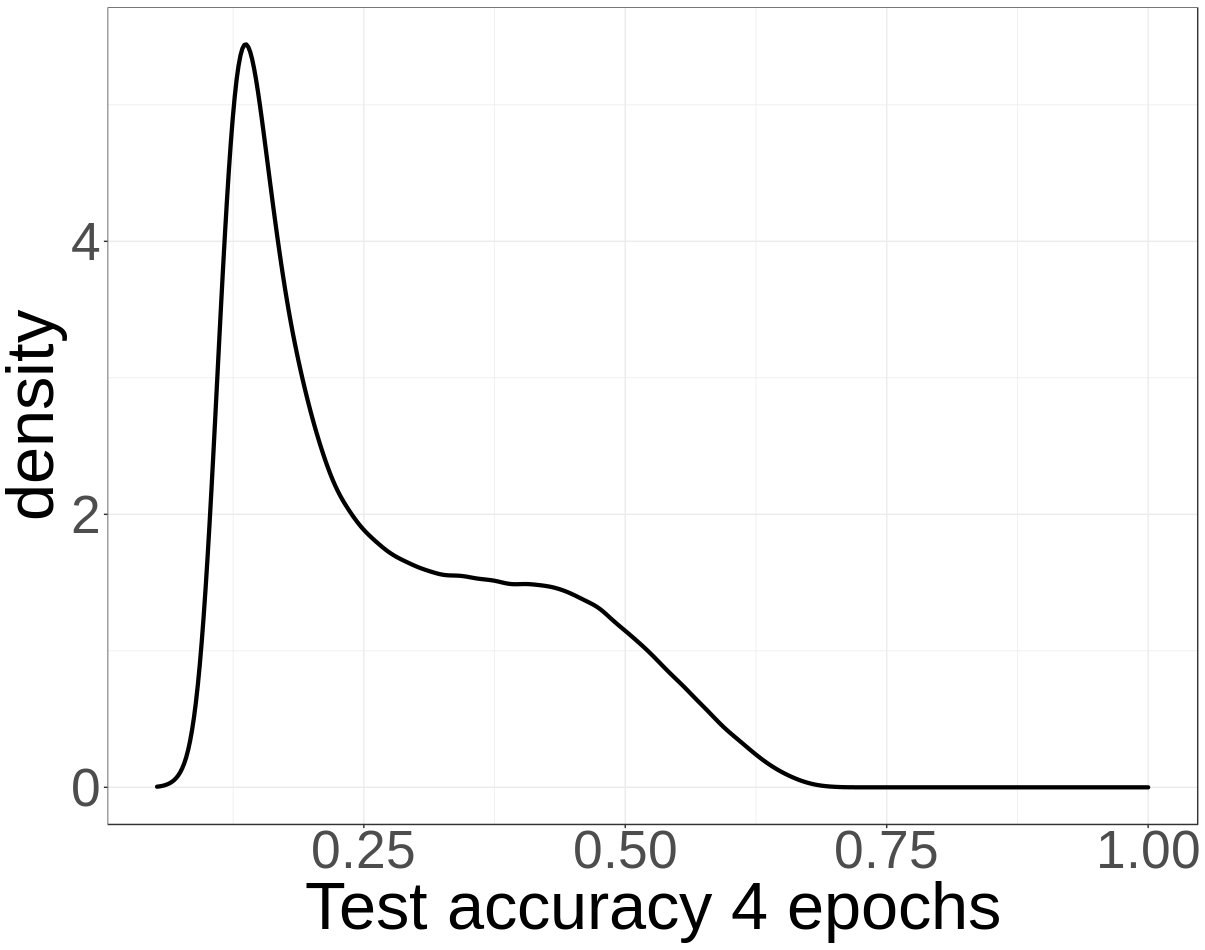}\hfill
  \includegraphics[width=.45\linewidth]{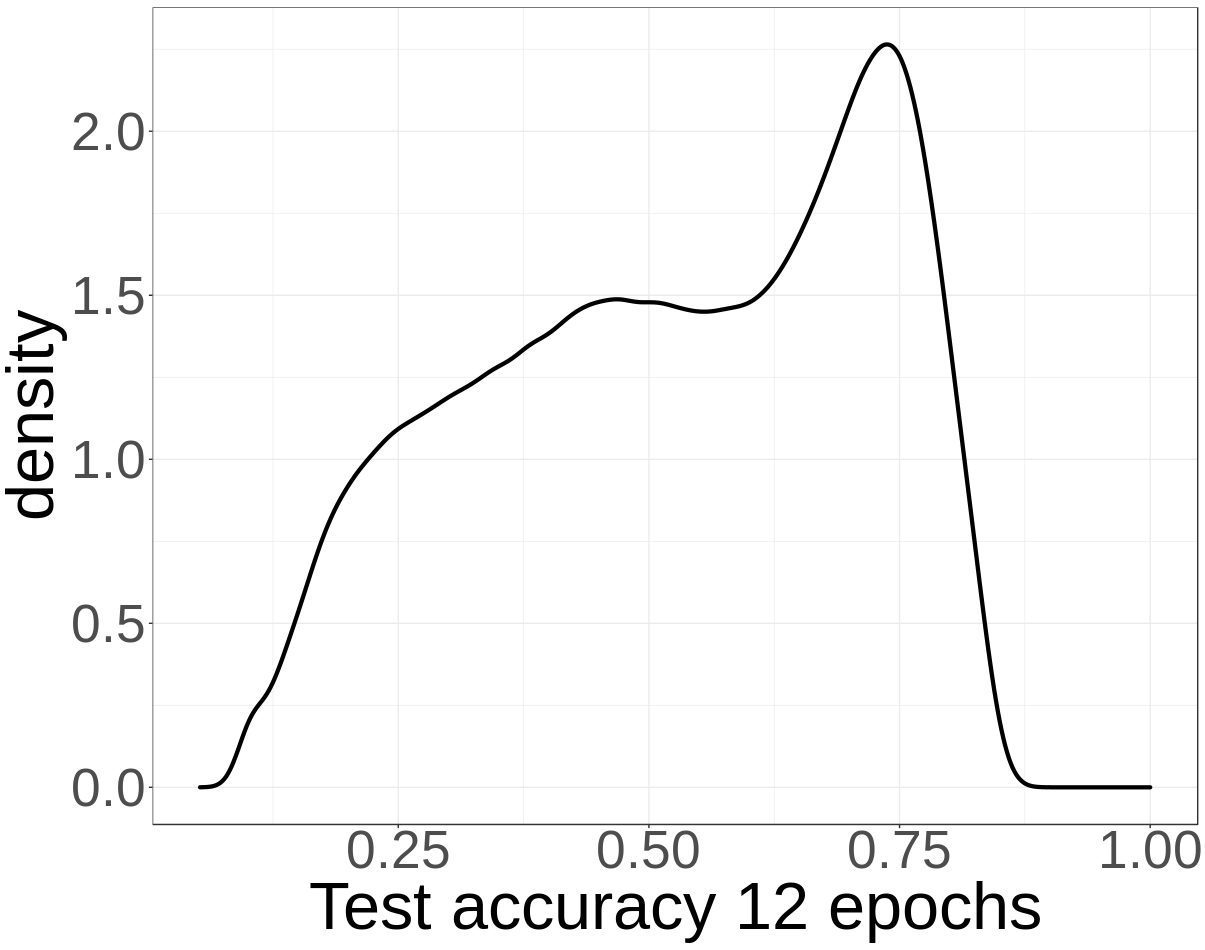}\hfill
  \end{subfigure}\par\medskip
  \begin{subfigure}{\linewidth}
  \includegraphics[width=.45\linewidth]{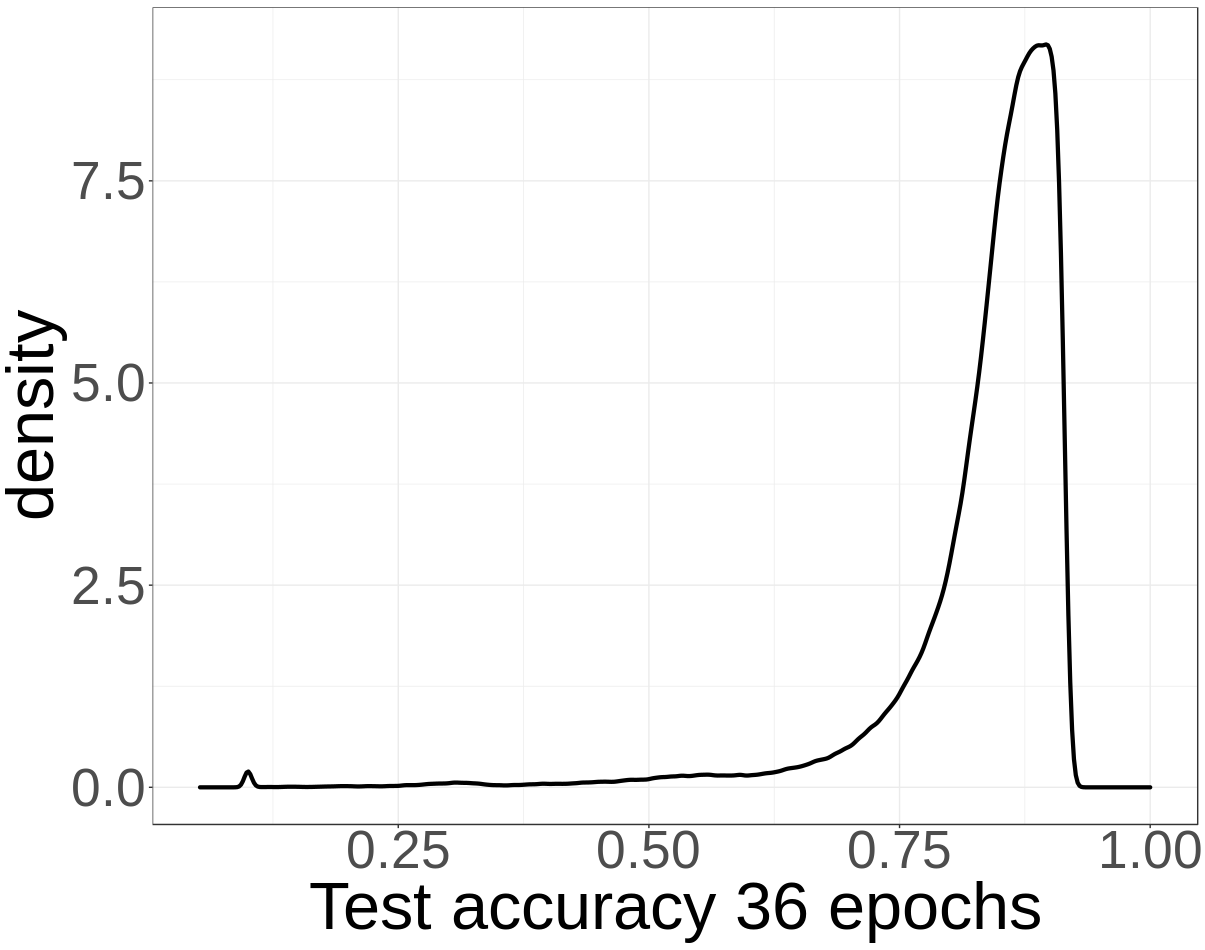}\hfill
  \includegraphics[width=.45\linewidth]{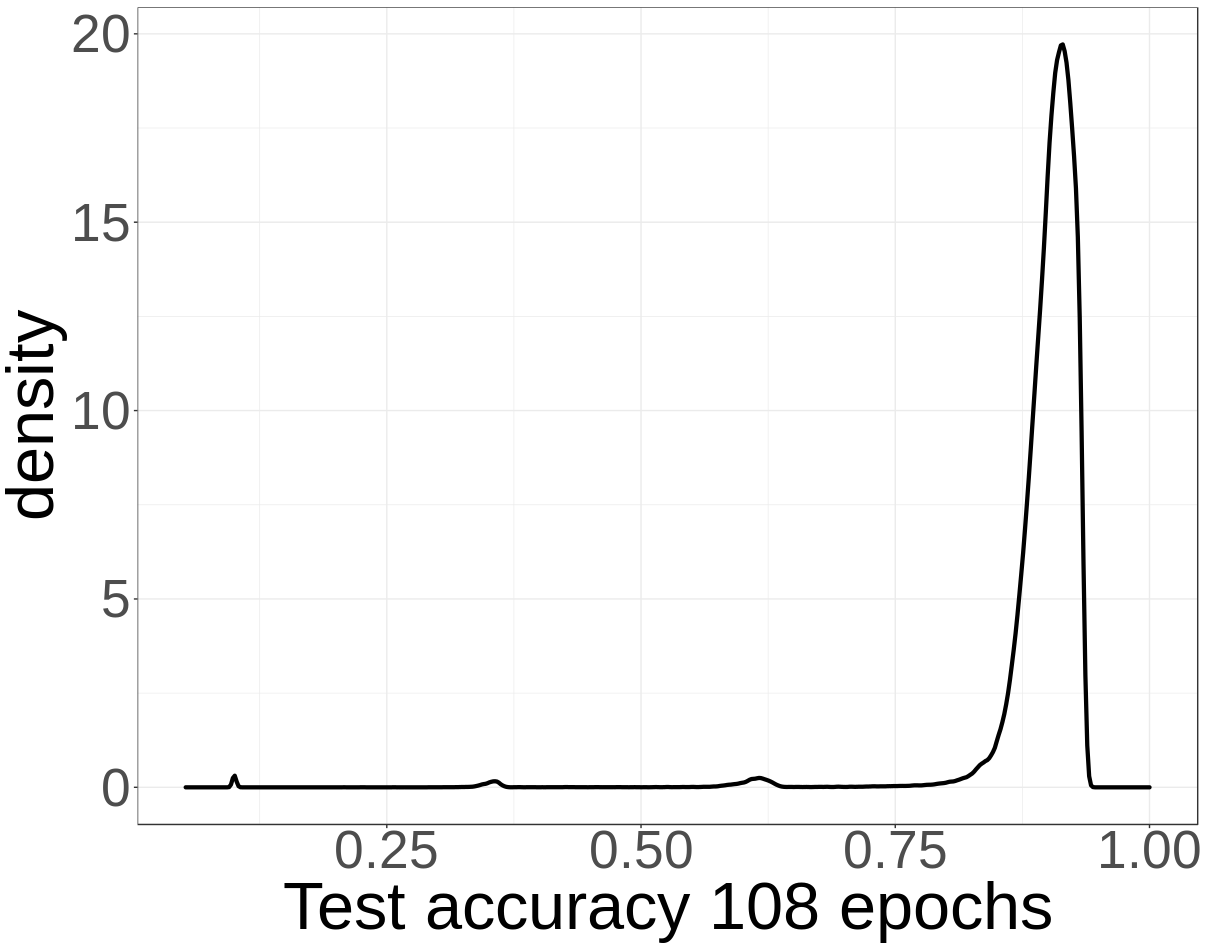}\hfill
  \end{subfigure}
  \caption{Density of fitness for various training budgets on CIFAR-10.}
  \label{fig:c10-pdf-test-acc}
\end{figure}

\begin{figure}[h]
  \begin{subfigure}{\linewidth}
  \includegraphics[width=.45\linewidth]{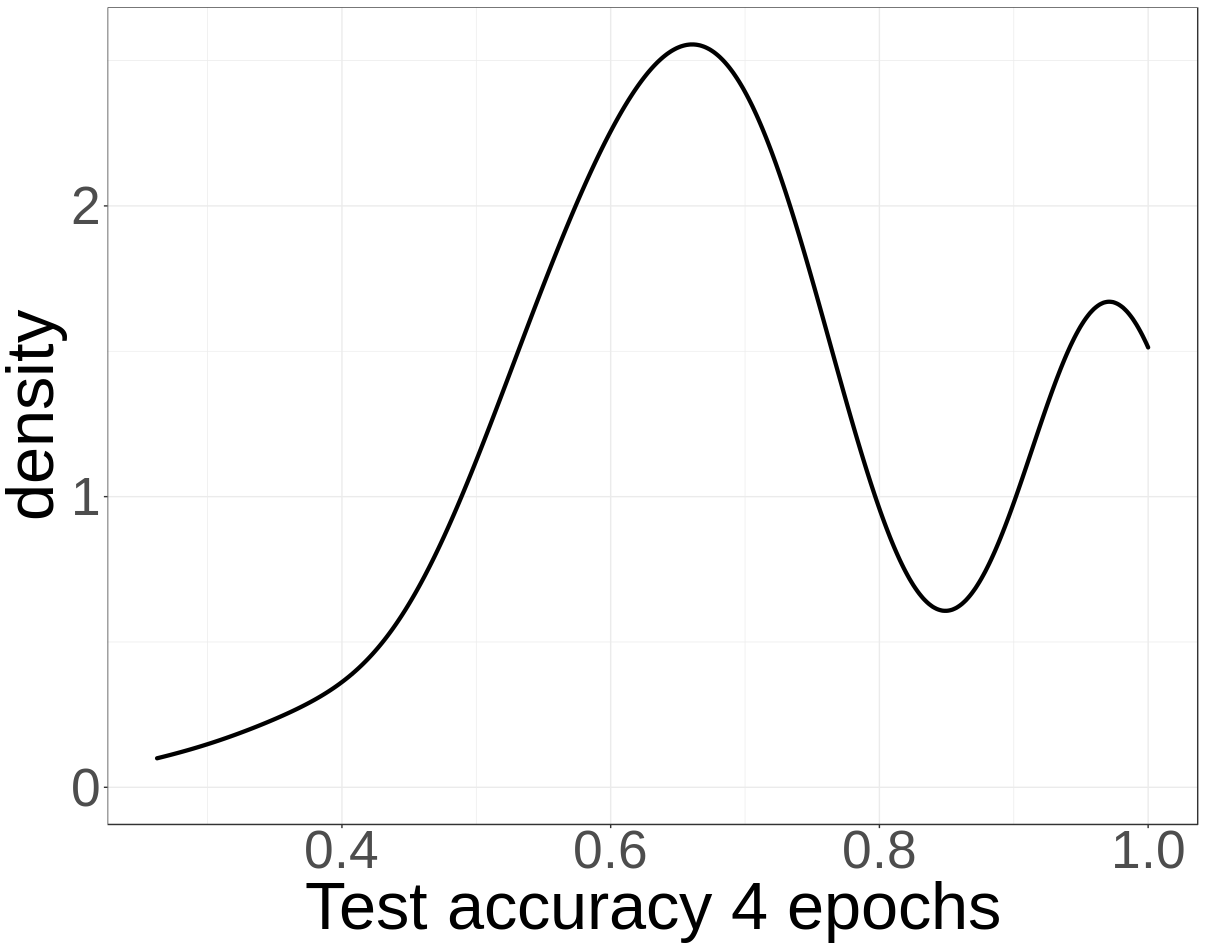}\hfill
  \includegraphics[width=.45\linewidth]{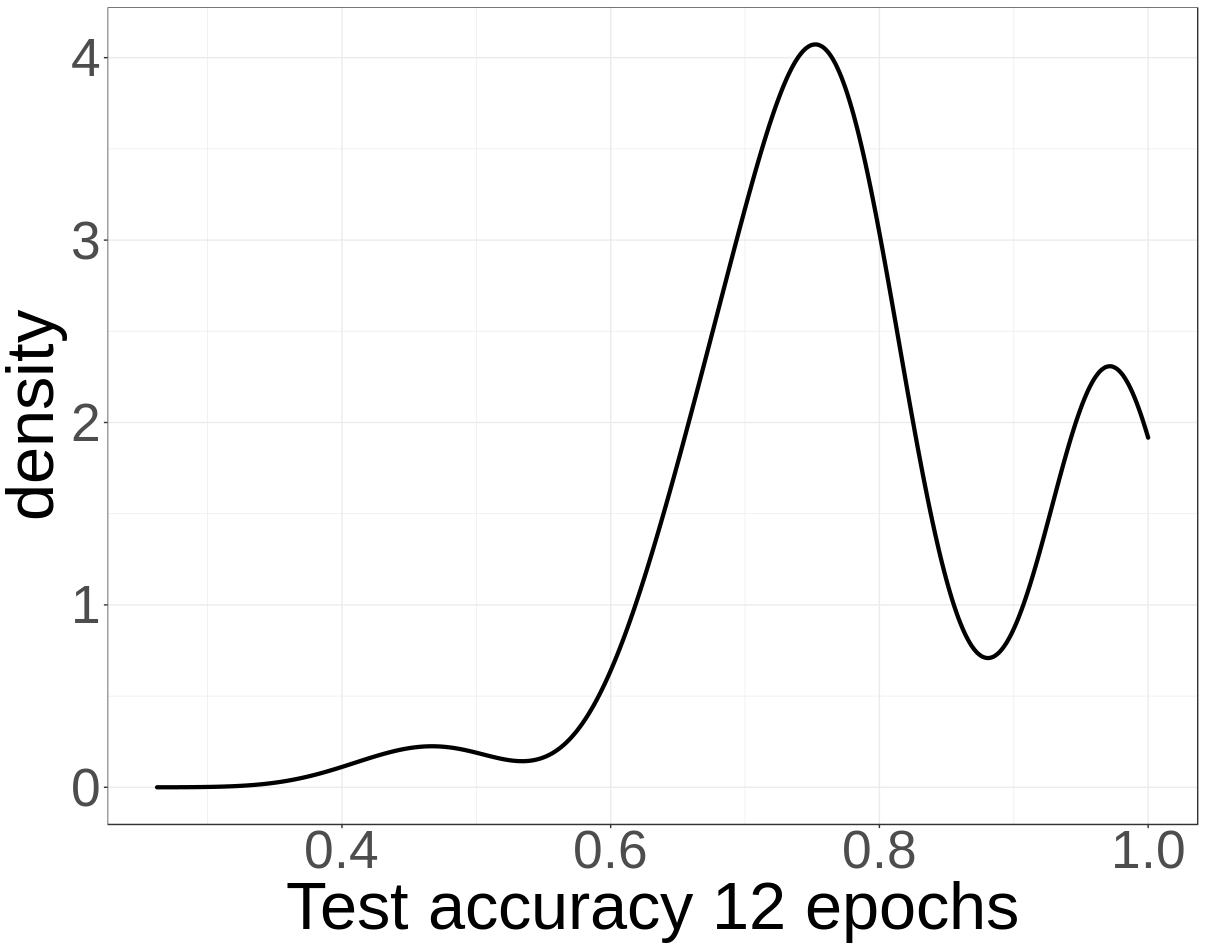}\hfill
  \end{subfigure}\par\medskip
  \begin{subfigure}{\linewidth}
  \includegraphics[width=.45\linewidth]{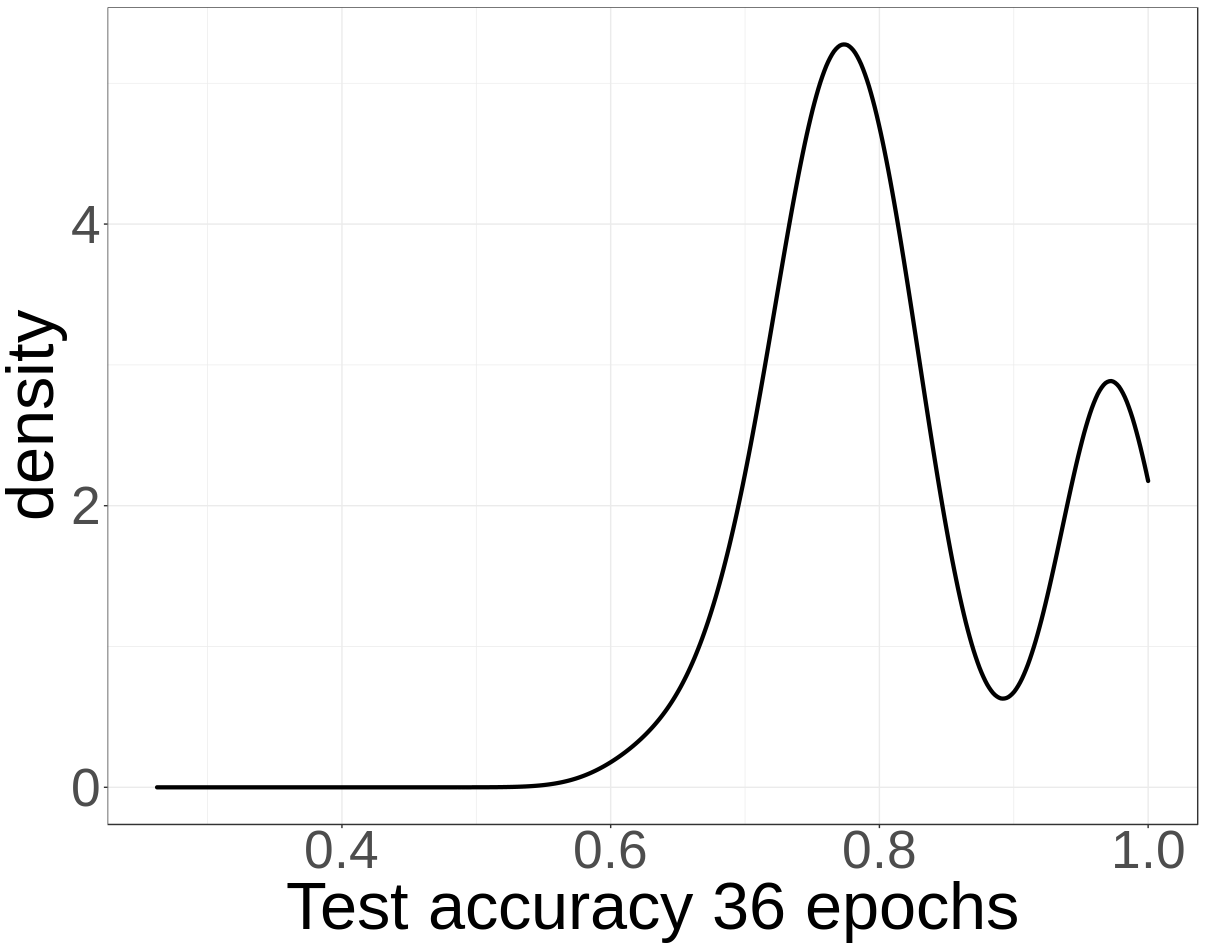}\hfill
  \includegraphics[width=.45\linewidth]{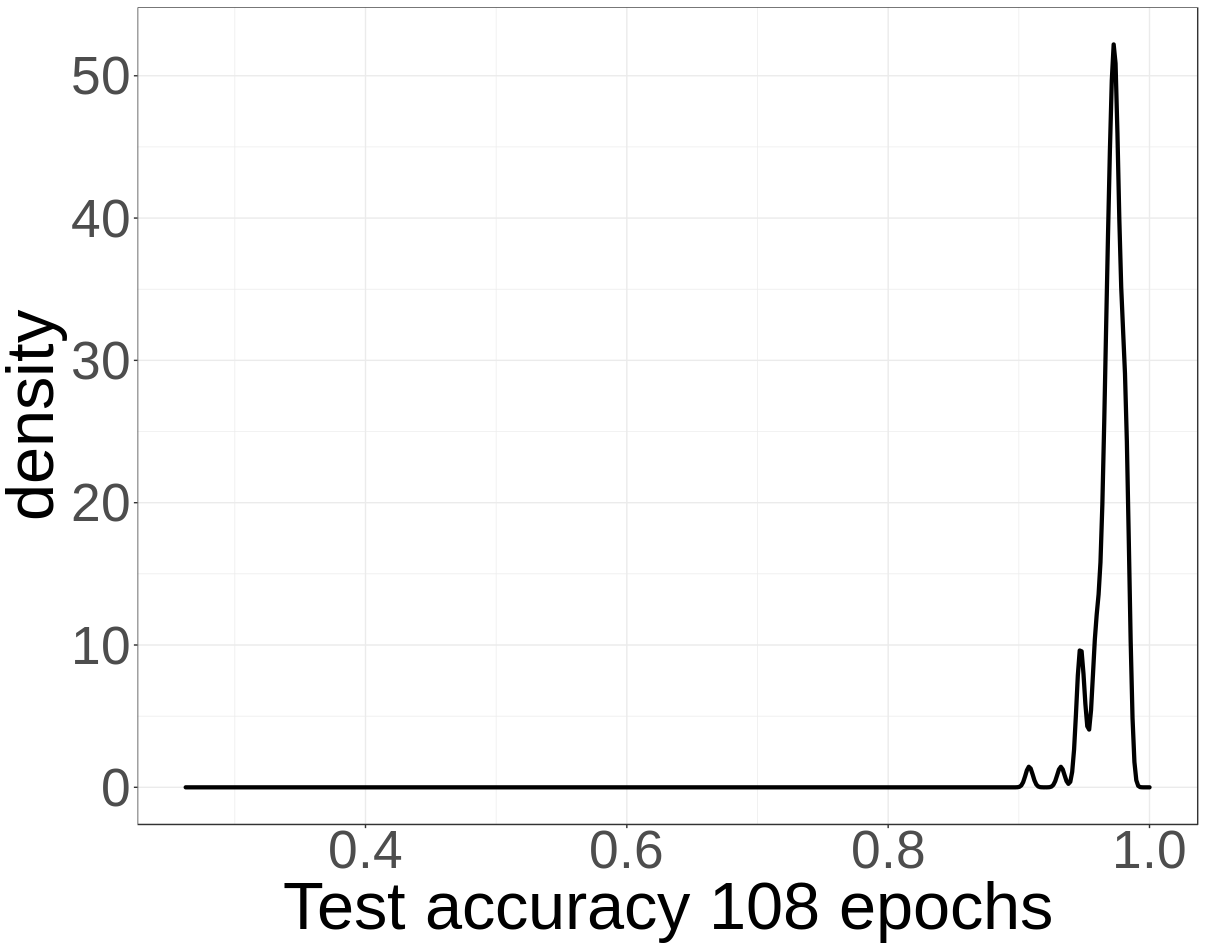}\hfill
  \end{subfigure}
  \caption{Density of fitness for various training budgets on So2Sat LCZ42. }
  \label{fig:lcz42-pdf-test-acc}
\end{figure}

In the context of NAS, 
the density of fitness measures the potential of a search space in fitting a given task.
Figures \ref{fig:c10-pdf-test-acc} and \ref{fig:lcz42-pdf-test-acc} 
show the density of fitness in test for various training budgets for CIFAR-10 and So2Sat LCZ42, respectively.

In both cases, the longer the training, the closer is the density to 1, the highest (and best) possible fitness value,
with most fitness above 75\% after 36 epochs. 
Thus, when trained long enough ($\geq$36 epochs), 
most solutions of the search space show very good fitting capacity on both data sets. 
Therefore, from the search perspective, 
there is a high chance of retrieving a good solution from~$\Omega$ on both domains, 
regardless of the ability of the NAS algorithm itself.



\begin{figure}[ht]
    \centering
    \includegraphics[width=0.95\columnwidth]{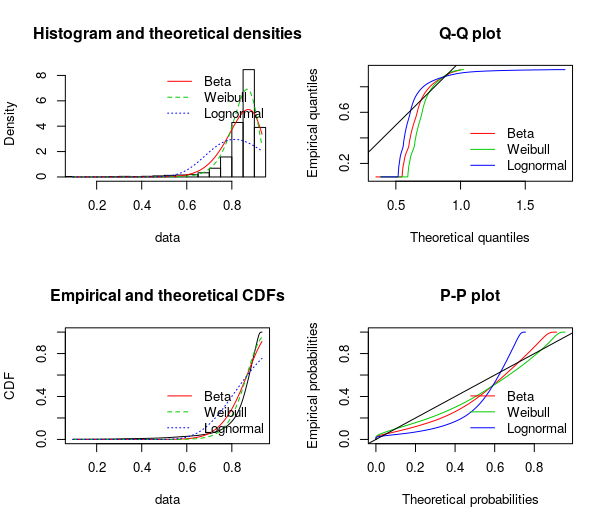}
    \caption{Comparison of empirical and theoretical densities of fitness on CIFAR-10, after 36 epochs of training.}
    \label{fig:c10-pdf-fit-36}
\end{figure}

In addition to measuring empirical densities of fitness, 
we investigate how they compare to theoretical distributions.  
Figure~\ref{fig:c10-pdf-fit-36} provides results for such comparison on CIFAR-10. 
The theoretical distributions are the Beta (red), Weibul (green) and Lognormal (blue).
The top-left and bottom-left hand figures show the associated histograms and  plots of cumulative density functions (CDFs) of fitness.
The top-right and bottom-right hand figures show the quantile-quantile (QQ) and percentile-percentile (PP) plots.

Table~\ref{tab:pdf-fitting-error-c10} presents the fitting error for each theoretical distributions.
The fitness is observed in validation after 36 epochs of training.
On CIFAR-10, PDF fitting is feasible and in favor of the Weibul distribution, 
which is close or on par with the empirical distribution of accuracy.
Indeed, in terms of fitting error the Weibul distribution generates the highest likelyhood and lowest AIC and BIC error values.

\begin{table}[h!]
 \centering 
\scriptsize
\begin{tabular}{|l|l|l|l|l|}\hline 
\textit{Error Metric / Function} &\textit{Beta} & \textit{Weibull} & \textit{LogNormal}\\
\hline
Likelihood & 5007293   & \textbf{534817} & 250868\\ 
\hline
AIC & -1001455  & \textbf{-1069630}  &  -501732 \\ 
\hline
BIC &-1001433   & \textbf{-1069608}  &  -501710\\ 
\hline
\end{tabular}
\caption{PDF fitting error on CIFAR-10} 
\label{tab:pdf-fitting-error-c10}
\end{table}

\begin{figure}[ht]
    \centering
    \includegraphics[width=0.9\columnwidth]{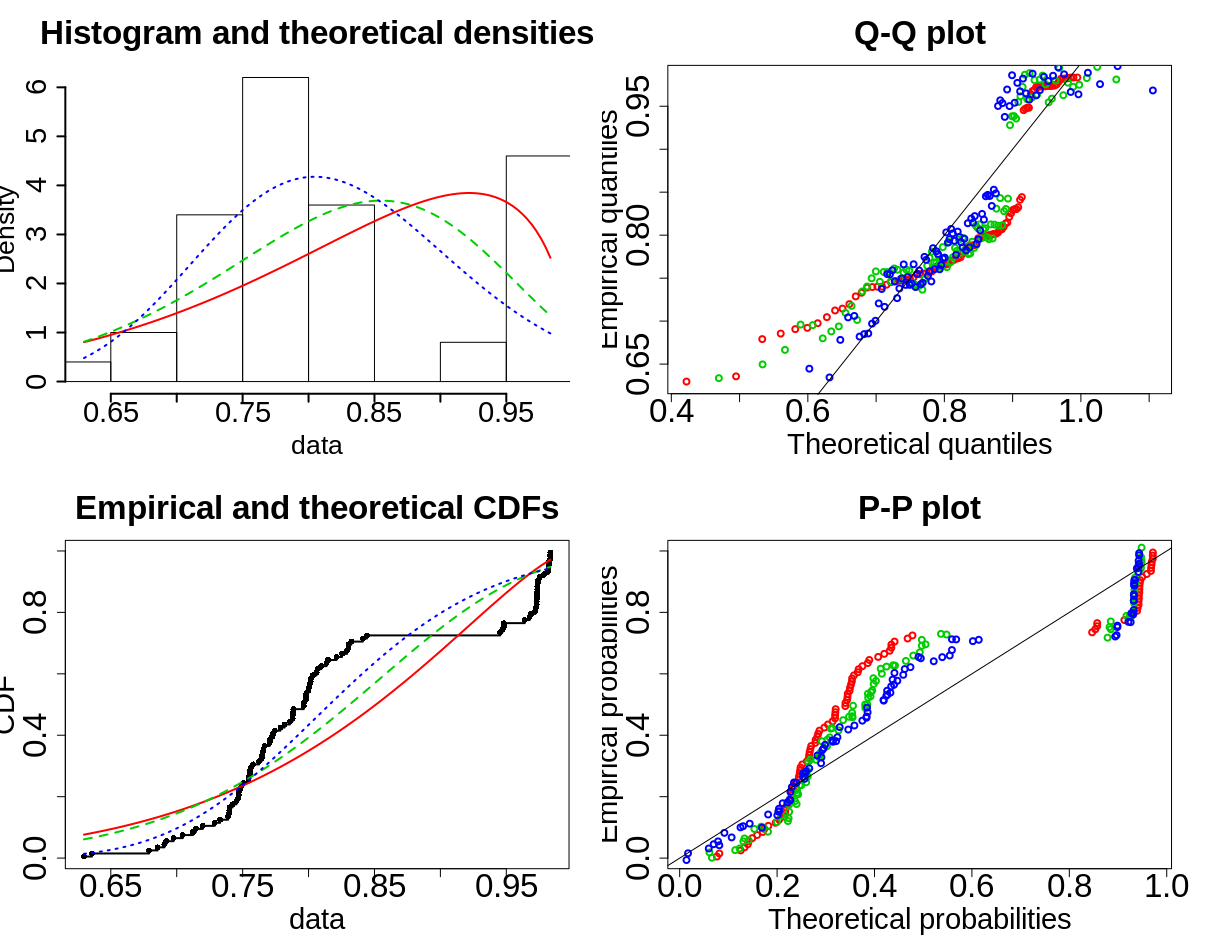}
    \caption{Comparison of empirical and theoretical densities of fitness on So2Sat LCZ42, after 36 epochs of training.}
    \label{fig:lcz42-pdf-fit-36}
\end{figure}

Figure~\ref{fig:lcz42-pdf-fit-36} and Table~\ref{tab:pdf-fitting-error-lcz42} present the results on So2Sat LCZ42.
In this case, the LogNormal distribution fits better the challenging multi-modal empirical distribution.
The other theoretical distributions fail to capture its second modality.

\begin{table}[h!]
 \centering 
\scriptsize
\begin{tabular}{|l|l|l|l|l|}\hline 
\textit{Error Metric / Function} &\textit{Beta} & \textit{Weibull} & \textit{LogNormal}\\
\hline
Likelihood &84.47 & 83.32 & \textbf{92.24}\\ 
\hline
AIC & -164.91 & -162.64  & \textbf{-180.48} \\ 
\hline
BIC &-159.73  & -157.43 & \textbf{-175.27}\\ 
\hline
\end{tabular}
\caption{PDF fitting error on So2Sat LCZ42} 
\label{tab:pdf-fitting-error-lcz42}
\end{table}

To summarize, the evaluated search space provides with high fitness values already after 36 epochs of training, on both datasets.
Results also indicate the possibility to model the distribution of fitness of~$\Omega$ for both datasets.

\subsection{Fitness Distance Correlation}

\begin{figure}[h]
    \centering
    \includegraphics[width=0.9\columnwidth]{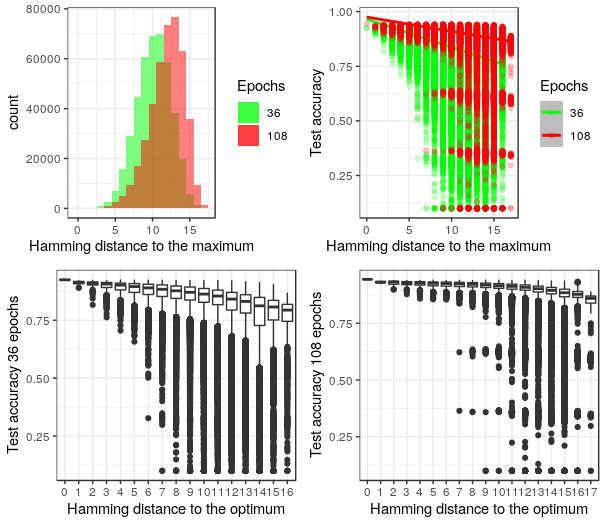}
    \caption{Fitness distance correlation on CIFAR10.
    }
    \label{fig:c10-test-fdc}
\end{figure}

\begin{figure}[h]
  \centering
  \includegraphics[width=.45\linewidth]{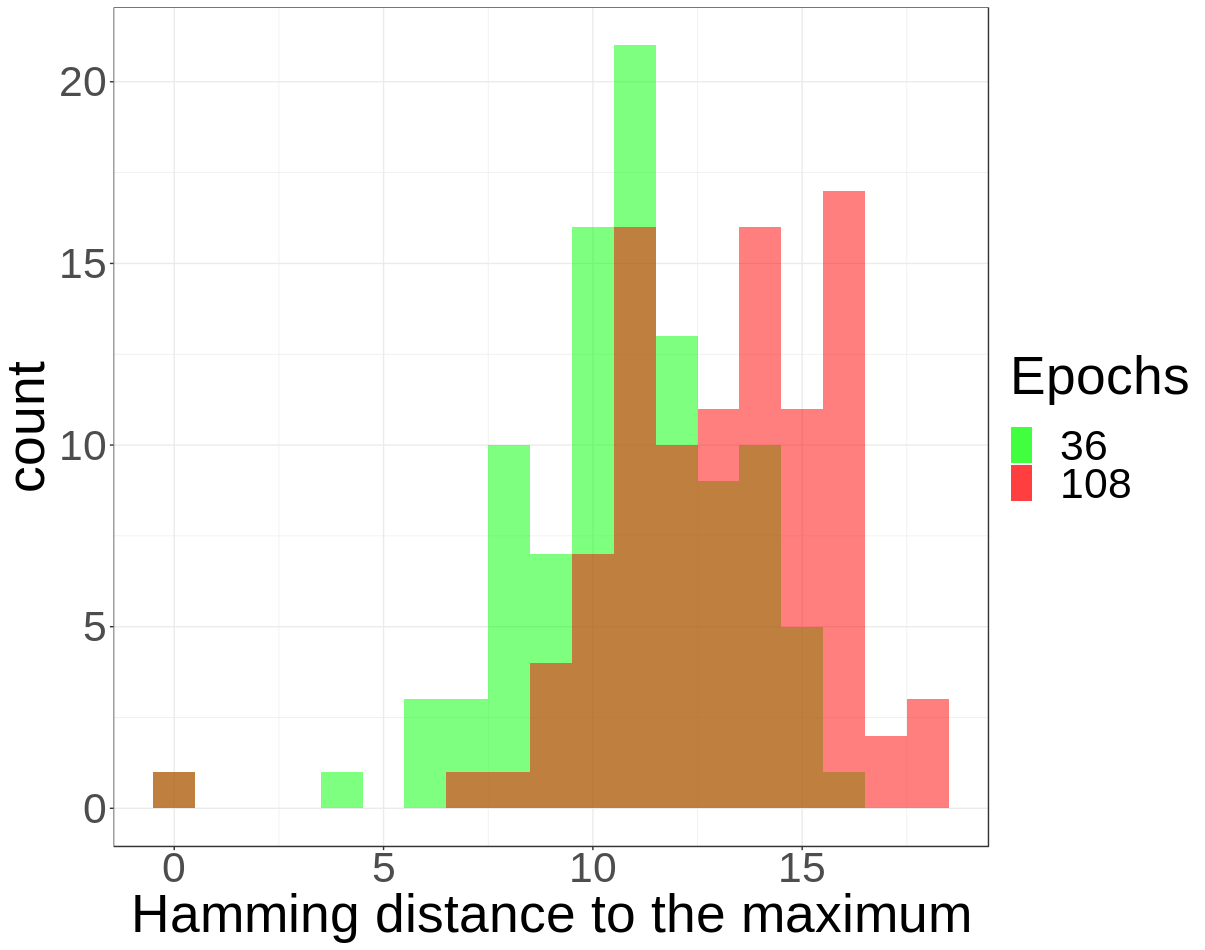}\hfill
  \includegraphics[width=.45\linewidth]{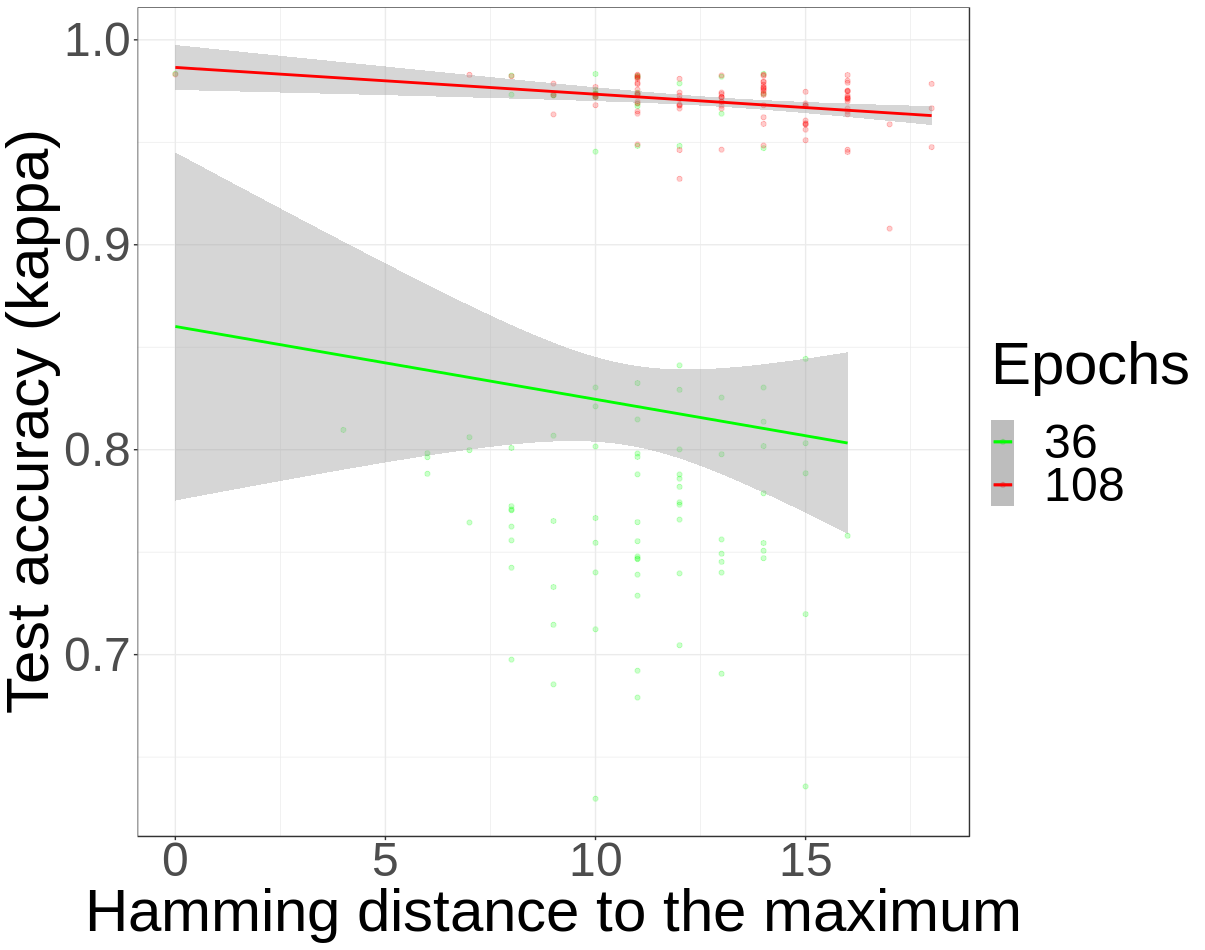}\hfill
 \centering
\includegraphics[width=.45\linewidth]{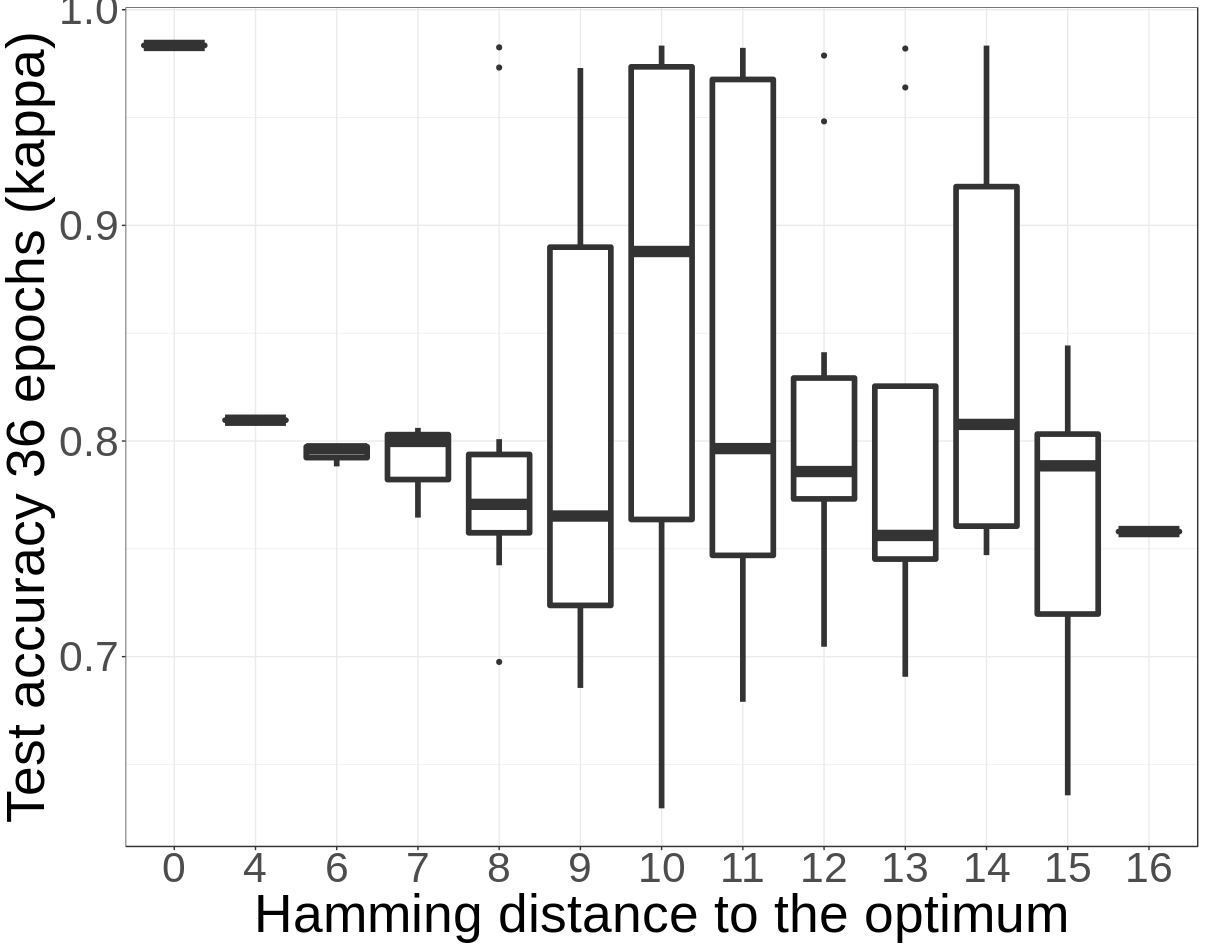}\hfill
\includegraphics[width=.45\linewidth]{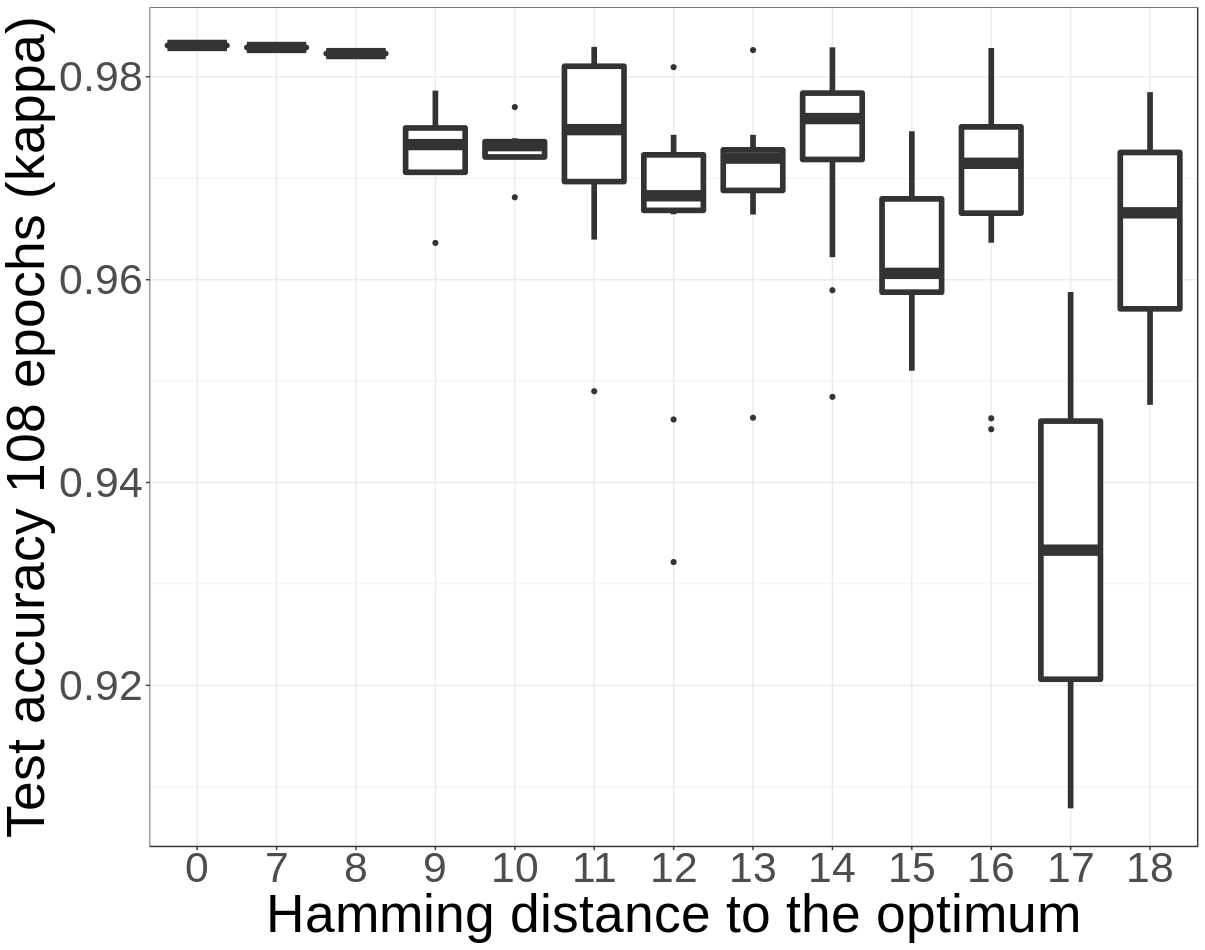}\hfill
 \caption{Fitness Distance Correlation on So2Sat LCZ42.}
 \label{fig:lcz42-test-fdc}
\end{figure}

%
Figures~\ref{fig:c10-test-fdc} and \ref{fig:lcz42-test-fdc} shows the FDC for CIFAR-10 and So2Sat LCZ42, respectively.
The top-left plot shows the histogram of the Hamming distance to the global optimum.
The top-right plot compares the FDC (linear fit) with the trained models (36 or 108 epochs).
The bottom-left and right figures show the individual FDC (36 and 108 epochs of training).
In all cases, the fitness is measured on the test set.

The histograms of distance to the optimum show, for both datasets, distributions appearing to be uni-modal and bell-shaped. 
They are centered around distance $d_{hamming}=10$ and covering a wide range of distances.
These results suggest that the models sampled on So2Sat LCZ42 are diverse and representative (of the search space). 
It also establishes a common ground for comparison of the FDC on both problems.

Regarding CIFAR-10, 
after 36 epochs of training, most solutions show a good fitness, i.e., above 70\%, on the whole spectrum of distances.
Also, a consistent increase in fitness (close to 1.66\%) per unit of distance travelled towards the optimum is observed. 
Moreover, aside from the outliers, we notice the regrouping of poorer performers at~$d_{hamming}=14$, 
creating a valley-like shape in the landscape.  
After 108 epochs of training, the increase of fitness when approaching the optimum diminishes, indicating a flat landscape or \emph{plateau}.  

Regarding So2Sat, 
most of the solutions perform well (above 60\% of fitness) after 36 epochs, regardless of their proximity to the optimum.
The non-consistent increase in fitness when approaching the optimum 
suggest high ruggedness and the existence of several local optima (e.g., $d_{hamming}=10,11,14$). 
In the long run the landscape becomes flat, with most solutions exhibiting high fitness (above 92\%).

To summarize, 
the FDC of both problems show a landscape improving towards high fitness as the training budget increases, and with a plateau-like shape.
Despite an apparent higher~ruggedness, NAS trajectories could benefit with higher gain from training only for 36 epochs. Therefore, we decided to focus the remainder of this study in the more challenging scenario of a training limited to 36 epochs.

  %
  %

\subsection{Ruggedness}

\begin{figure}[h]
  \centering
  \includegraphics[width=0.75\columnwidth]{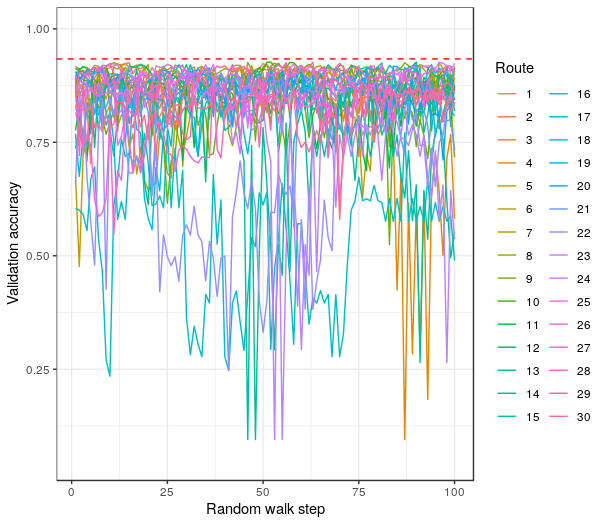}
  \caption{Random walk routes after 36 epochs on CIFAR-10.}
  \label{fig:c10-rand-walk-routes}
\end{figure}

To analyze the ruggedness, we study the behavior of random walks in~$\Omega$. 
Figure~\ref{fig:c10-rand-walk-routes} 
shows evaluations of thirty (30) independent random walk routes on CIFAR-10. 
Each route is composed of one hundred (100) steps, starting from a randomly sampled configuration.

\begin{figure}[h]
  \centering
  \includegraphics[width=0.75\columnwidth]{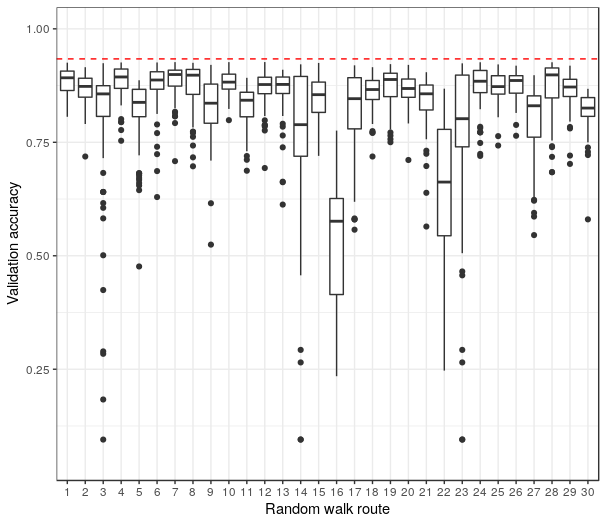}
  \caption{Distribution of the fitness of the random walks on CIFAR-10 (36 epochs training).}
  \label{fig:c10-rand-walk-boxplot}
\end{figure}

Figure~\ref{fig:c10-rand-walk-boxplot} shows the fitness distribution for each route.
Overall, there is a high fluctuation in fitness from one route to the other.
Most exhibit a fitness above 75\%, even though some have distributions in the lower end, 
suggesting walks being stuck near local minima. For example, route 22.

Particularly, two routes stand out given their distributions of fitness. 
Route 1 (R1) displays a median (med=0.89) in the higher end, with little variance (std=0.03),
while route 16 (R16) has the lowest median of all (med=0.58), with a widespread distribution (std=0.14). 

\begin{figure}[h]
  \centering
  \begin{subfigure}{\linewidth}
  \includegraphics[width=0.85\columnwidth]{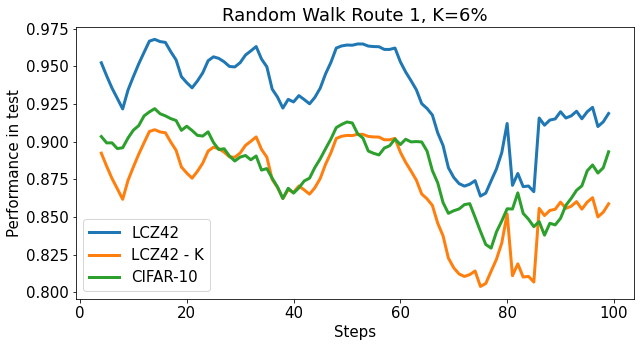}
  \label{fig:lcz42-r1}
  \end{subfigure}
  \begin{subfigure}{\linewidth}
  \includegraphics[width=0.85\columnwidth]{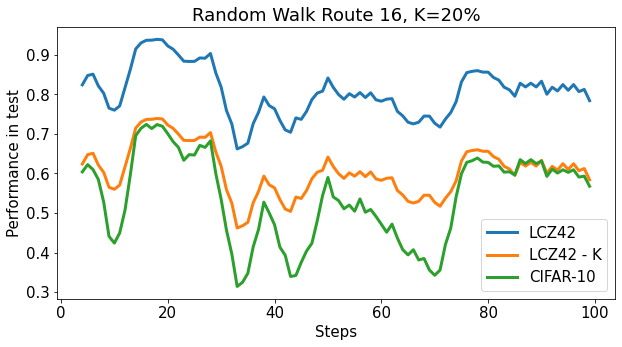}
  \label{fig:lcz42-r16}
  \end{subfigure}
  \caption{Random walk routes 1 and 16 on CIFAR-10 and So2Sat LCZ42. Fitness in test for models trained 36 epochs.}
  \label{fig:c10-lcz42-r1-r16}
\end{figure}

As a ground of comparison, we also evaluate those two extremes on So2Sat LCZ42.
Results for both datasets are presented in Figure~\ref{fig:c10-lcz42-r1-r16}. 
All curves result from a processing with a moving average of five (5) steps.

For R1, we observe distributions that are centered around high values with little variance.
Both paths show several identical bumps and curvatures.
Subtracting a constant value $K=6\%$ to the fitness on So2Sat LCZ42 results in a curve close to R1 on CIFAR-10.
Similarly, for R16, the route on So2Sat LCZ42 shows a curvature similar to CIFAR-10.
Removing a constant $K=20\%$ from the fitness on So2Sat LCZ42 results in a curve closely fitting R16 on CIFAR-10.
Therefore, the results  suggest a similar ruggedness on both datasets,
for the areas of the landscape being explored with R1 and R16. 
Additionally, it is an indication of a common imprint of the search space~$\Omega$ on trajectories of NAS, across datasets.
So far, we identify a variability~$K$ in fitness, such as:~$ 6\% \leq K \leq \%20$.

Summarizing, the variability (fitness) observed in the random walks suggests that there are areas in $\Omega$ with higher (or lower) fitness potential. 
The comparison of a few routes show similar trajectories and ruggedness on both datasets.

\subsection{Local Optima}

To complement the analysis of \textit{ruggedness},
we look to quantify the number of \textit{local optima} in~$\Omega$.
Table~\ref{tab:local-optima-estimation} presents the results of the estimation on CIFAR-10.
The fitness consist in the accuracy in test after 36 epochs of training. 
The estimation is obtained by executing Algorithm~\ref{algo:local-optima-enumaration} for $T=9$ trials.
For each trial~$i$, we perform $M=200$ runs of \emph{best improvement local search} in order to measure the index of the first occurrence of duplicates~$k_i$. Note that because of the ascending nature of the BILS algorithm, it leads to local maxima.

\begin{table}[h!]
 \centering 
\scriptsize
\begin{tabular}{|l|l|l|l|l|}\hline 
\textit{Trial} &\textit{Avg. Step} & \textit{Avg. Improvement(\%)} & \textit{First Repeat k} & \textit{Cardinal}\\
\hline
1 & 2.90 & 4.53 & 94 & 6373 \\
\hline
2 & 2.89 & 5.48 &57 & 2343 \\
\hline
3 & 2.78 & 4.76 &26 & 487 \\
\hline
4 & 3.02 & 5.55 & 58 & 2426 \\
\hline
5 & 2.86 & 5.06 & 38 & 1041 \\
\hline
6 & 3.01 & 4.5 & 195 & 27429 \\
\hline
7 & 2.83 & 5.77 & 52 & 1950 \\
\hline
8 & 2.80 & 4.78 & 129 & 12003 \\
\hline
9 & 2.70 & 4.79 & 197 & 27994\\
\hline
\hline
\textbf{Summary} & \textbf{2.86} & \textbf{5.03} & \textbf{94} & \textbf{6373}\\
\hline
\end{tabular}
\caption{Enumeration of \textit{local optima} (maxima) in CIFAR-10 via the \textit{birthday problem} statistics. 
} 
\label{tab:local-optima-estimation}
\end{table}

On average, the length of the step is 2.86, with an improvement in the fitness of 5.03\%. 
Over the whole experiment, the average index of repeat for a budget of $M$ runs is $k_{mean}=94$.
Using Equation~\ref{eq:card_opt} and~$k_{mean}$, we estimate $N_{CIFAR-10}=6373$ 
local optima (maxima) on $\Omega$ (CIFAR-10).
Besides, we report six (6) additional failing runs with no duplicates found within $M$ runs.

Next, we look to approximate this value on So2Sat LCZ42 with only a few evaluations.
Assuming a continuity in fitness for neighboring solutions in~$\Omega$, i.e., locality, 
we derive the number of local maxima~$N_{proxi}$ 
by counting solutions with higher fitness 
than their~$N_{nei}=10$ nearest neighbors.   
Using such approximation for the $N_{samples}=100$ identical samples on both datasets,
we derive a ratio of enumeration using:
\begin{equation}
r_{optima}=\frac{N_{proxi-LCZ42}}{N_{proxi-CIFAR-10}}
\label{eq:proxi_card_opt}
\end{equation}
Then, we derive final the enumeration on So2Sat LCZ42 using: 
\begin{equation}
 N_{LCZ42}=r_{optima} * N_{CIFAR-10}
\label{eq:card_opt_LCZ42}
\end{equation}
where, ~$r_{optima}=75\%$ is the obtained ratio, and ~$N_{CIFAR-10}=6373$ the ground truth approximation on CIFAR-10. 
This leads us to ~$N_{LCZ42}=11153$, the approximation of the enumeration on LCZ-42. 

Overall, results indicate approximately~$75\%$ more local optima on the landscape of LCZ-42 than on CIFAR-10.

\subsection{Persistence}
\begin{figure}[h]
  \begin{subfigure}{\linewidth}
  \centering
  \includegraphics[width=.85\linewidth]{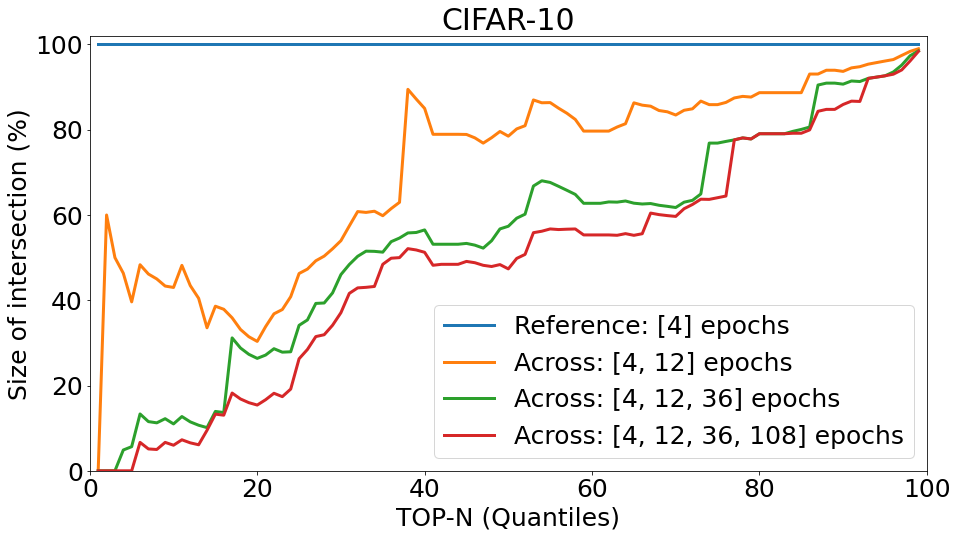}\hfill
  \caption{}
  \end{subfigure}
  \begin{subfigure}{\linewidth}
  \centering
 \includegraphics[width=.85\linewidth]{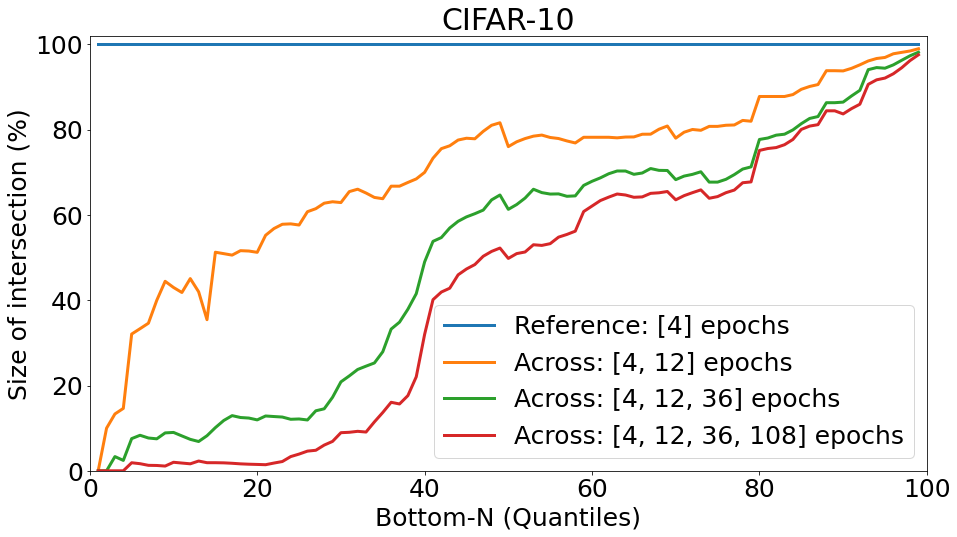}\hfill
  \caption{}
  \end{subfigure}
  \caption{Positive and Negative Persistence across training regimes in CIFAR-10.}
  \label{fig:c10-persistence}
\end{figure}

Also, we analyzed the samples from the perspective of \emph{persistence} in their rank over time.
Figure~\ref{fig:c10-persistence} depicts results of \emph{persistence} on CIFAR-10, considering~$N_{samples}=1000$ samples and their fitness in test. 
The top and bottom plots show the positive and negative persistence, respectively.
For each plot, the blue curve represent the reference population, i.e., the set of models at a given $Rank-N$ (Nth percentile) considering the fitness after 4 epochs of training.
The yellow curve stands for the share of these models maintaining the same $Rank-N$ after 12 epochs.
The green and red curves are for the remaining models also maintaining the same $Rank-N$ after 36 and 108 epochs of training.

Regarding CIFAR-10, 
the chance of remaining $Top-N\%$ until 12 epochs is rather high for all N (all above 30\%), 
despite an important drop with N below 40.
We notice that the chance of remaining $Top-N\%$ performer until 36 epochs and until 108 epochs,
are almost overlapping, with a consistent decrease as N increases.
Similarly, the chance of remaining $Bottom-N\%$ performer until 36 and until 108 epochs are tied,
even tough for N below 20, chances tend to zero.
If a model is $Top-N\%$ (or $Bottom-N\%$) performer until 36 epochs,
it will most likely remain $Top-N\%$ (or $Bottom-N\%$) performer until 108 epochs.

\begin{figure}[h]
  \begin{subfigure}{\linewidth}
  \centering
  \includegraphics[width=.85\linewidth]{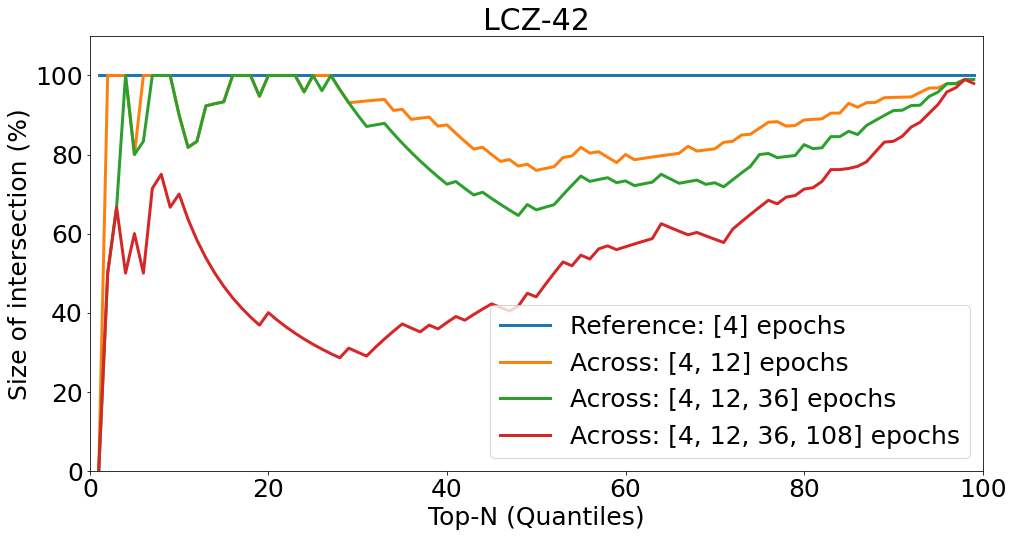}\hfill
  \caption{}
  \end{subfigure}
  \begin{subfigure}{\linewidth}
  \centering
  \includegraphics[width=.85\linewidth]{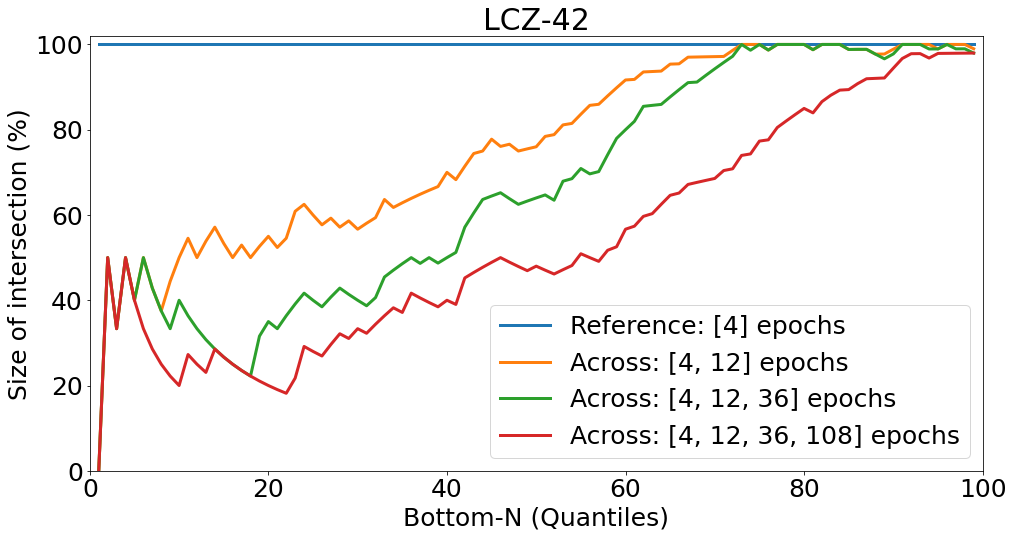}\hfill
  \caption{}
  \end{subfigure}
  \caption{Positive and Negative Persistence across training regimes in So2Sat LCZ42.}
  \label{fig:lcz42-persistence}
\end{figure}

Figure~\ref{fig:lcz42-persistence} introduces the \emph{persistence} for So2Sat LCZ42. 
In this case, the positive persistence remains quite high (above 25\% for all N).
For the elite performers (Q1), we notice a high peak in persistence as N drops below 30.
On the other hand, the negative persistence displays a steady drop until N equals 20.
Results on this problem indicate in general higher positive and negative persistence, 
in particular for the elite N-performers, solidifying their rank over time.

Overall, 
the positive persistence in both problems evolves similarly, and it is noticeable even for ranks below the first quartile (Q1).  
On the other hand, the negative \emph{persistence} on LCZ-42 remains consistently higher.

\subsection{Fitness Landscape Footprints}
\noindent
Finally, we present the~\emph{fitness footprint} for both datasets.
Figure~\ref{fig:footprints} depicts the \emph{footprint} of CIFAR-10 in blue, and So2Sat LCZ42 in orange. 
Except for \emph{persistence}, the fitness (test) is considered for 36 epochs, 
as the landscapes appear more challenging at this stage. 
The main takeaways are:
\begin{itemize}
    \item ~\emph{Overall fitness}: Both data sets show a similar overall fitness (around 83\%), 
    but So2Sat LCZ42 present a slightly higher deviation (10\%), suggesting more variability in the fitness of its search space.
    %
    \item ~\emph{Ruggedness}: The 10\% higher \emph{ruggedness} on CIFAR-10 (1.93 instead of 1.75) 
    is coherent with the qualitative assessment made of the random walks (Figure~\ref{fig:c10-lcz42-r1-r16}),  
    suggesting more fluctuations and difficulties to overcome by \emph{local search}-based algorithms on CIFAR-10 than on So2Sat LCZ42. 
    While this was assessed on 36 epochs, we expect it to hold true for longer training regimes. 
    \item ~\emph{Cardinal of optima}: With approximately 75\% more local optima on So2Sat LCZ42 (11153 versus 6373),
    the chances of being trapped in a non-optimal region of the space are higher on So2Sat LCZ42 than on CIFAR-10. 
    Then, solution \emph{diversity} should be considered when designing a 
    (\emph{local search}-based) algorithm to do NAS on So2Sat LCZ42 (i.e., exploration-exploitation). 
    %
    \item ~\emph{Positive persistence}: Elite models persist over time, i.e., the chances of finding a model that will be top-25\% performer in test from 4  until 108 epochs of training is 32\%. 
    Moreover, on So2Sat LCZ42, elite models (rank~$<25\%$) will remain on the top with higher probability (0.48).
    Therefore, spotting elite performers early could help to improve NAS performance.
    \item ~\emph{Negative persistence}: 
    Furthermore, we observe on So2Sat LCZ42 a higher chance of models remaining bottom-25\% performers across training time (28\% versus 6.4\%) and a larger~$AuC$ of \emph{persistence} for ranks below 25\%. 
    In other words, it is important to avoid poor performers early is more critical on So2Sat LCZ42, while it is not a concern on CIFAR-10.
\end{itemize}

Overall, we observe similarities in the characteristics of the ~\emph{footprints} of both problems.
Some key aspects such as the \emph{persistence} could be help improve \emph{local search} baselines on both problems.

\section{Discussion}

The following paragraphs provide a discussion of the results obtained.
First, we discuss results of each individual aspects of the \emph{fitness landscape analysis}.
Then we discuss the~\emph{footprint} as a way to summarize the analysis.

\emph{Density of fitness:} 
Our first experiments consisted in measuring the PDFs of fitness on both problems,
and attempting to fit them with various theoretical distributions.
In doing so, we notice similarities in the evolution of densities in both problems. 
Indeed, the PDFs tend to be uni-modal, very narrow distributions and close to~$1$, the longer the training.
In other words, most solution of the search space have high-capacity fitting both datasets, given enough training.
We consider this as a first indication that most NAS search strategies might perform well with 108 epochs of training,
regardless of their search ability (on the given problems).
Thus, when benchmarking NAS performances on both datasets,
we recommend the use of less training in order to better differentiate the algorithm ability.  

\emph{Fitness Distance Correlation:} 
Investigating the FDC enabled us to uncover the potential 
gain in fitness per travelled hamming distance for a NAS optimizer. 
The longer the training, the flatter the fitness landscapes, with high fitness across all the distances to the global optimum.
These results indicate that a NAS algorithm 
can expect higher gain per iteration with intermediate training budget, i.e., 36 epochs.
Indeed, the flat shape of the landscape after 108 epochs allows for a more limited margin in NAS \emph{trajectories}.

Moreover, the results hold for \emph{local search} based algorithms, as well as evolutionary computation-based approaches,
on the given search space and sample encoding.

\emph{Ruggedness:}
 Besides, the analysis of the random walks on both problems (see Figure \ref{fig:c10-lcz42-r1-r16}) for two different routes shows that the fitness on both datasets are just at a constant $K$ away, with a similar curvature.
 This suggests a strong influence from a common element of the landscapes: the search space~$\Omega$.

Additionally, the slightly larger \emph{ruggedness} coefficient on CIFAR-10 suggests slightly 
more fluctuations in local search trajectories for the dataset. 
However, because of the expensive nature of the experiments, we estimated the coefficient on LCZ-42 out of fewer walks (2 versus 30).
Future work could investigate a more in depth measurement and comparison, given a larger budget of evaluations.
This could help better understand how the locality in fitness differs from one landscape (dataset) to the other.

\emph{Local optima:} 
We proceeded to describe the landscapes by assessing the existence of \emph{local optima}. 
Our estimation indicated a larger number of local optima on So2Sat LCZ42 than on CIFAR-10.
This suggests adapting local search-based NAS Algorithms to this problem, in order to avoid getting \emph{stuck} 
in suboptimal areas of the landscape. 

Besides, we hereby comment some aspects of the enumeration on CIFAR-10, 
using the Algorithm~\ref{algo:local-optima-enumaration}.
We reported a few (6) failing attempts to recover collisions out of~$M=200$ runs. 
This suggests that for some seeds, the limited budget~$M$ of repetition was not enough, 
and involved a slight underestimation of the average index at collision~$K_{means}$.
A more accurate estimation could be obtained with a larger budget~$M$ of repetition.
In this work, we mainly aimed at providing an early approximation of this value with a restricted budget~$M$, 
following~\cite{Matthew99estimatingCardBigSS}.

\emph{Persistence:} 
Then,
we investigated the samples from the perspective of a proposed metric, the \emph{persistence}.
Results indicate a similar positive \emph{persistence} of 33\% on So2Sat LCZ42 and CIFAR-10.
This implies that chances are significant of finding good models early in their training, in both datasets.
From the perspective of search, this also suggest that the chances that an area of the search space remains 
\emph{fruitful} ($Top-Nth$ performers) are high.

On the other hand, we observed a higher negative  \emph{persistence} and greater \emph{areas under the curve} (positive and negative) on So2Sat LCZ42.
Overall, the high persistence (positive or negative) on So2Sat LCZ42 indicate good chances of finding early in training 
models, that will perform god (or bad) after longer training.
Suggesting a potential gain by spotting good and bad performers early (i.e., considering a short training regime).
From the perspective of the search, this also suggest that chances are quite high that a region of the search space remains 
\emph{fruitful} (with $Top-Nth$ performers) or \emph{bad} (with $Bottom-Nth$ performers) over training time.

Overall, results are in favor of performing search with a short training budget (at most 36 epochs),
as models might have a good \emph{persistence} in their fitness.

\emph{Fitness footprint:}
Finally, we introduced the \emph{footprint} as a way to summarize the previous aspects of a FLA.

Regarding So2Sat LCZ42, the~\emph{fitness footprint} 
tells us that we can expect good but 
slightly more variable fitness on the search space of NASBench-101. 
It also informs us that when deploying NAS on So2Sat LCZ42 there might be potential margin to gain and losses to avoid using simple heuristics. 
Indeed, a heuristic to spot elite models early in training could help filter the search space in order to focus on more fruitful areas of the search space. On the other hand, a heuristic to avoid poor performers early could 
help reduce the overall complexity by saving training time in selection. 
Another critical heuristic to have is one enabling diversity in \emph{local search} helping to avoid bad regions of the search space.    
As there is potentially more adjustment in terms of heuristics in order to avoid losses in fitness 
(larger cardinal of optima, negative persistence), 
So2Sat LCZ42 might be a slightly more challenging problem for NAS given the current search space. 

Regarding CIFAR-10, the~\emph{fitness footprint} also describes overall good and stable performances 
on the search space.
A larger~\textit{ruggedness} lets us anticipate more difficulties for deploying \emph{local search}. 
On the other hand, 
CIFAR-10 also exhibits potential gains to have by incorporating heuristics for spotting early elite models when deploying NAS algorithms. 

Overall, we observe similar characteristics in 
both~\emph{footprints}. Therefore, we may extrapolate from one problem to the other.

\section{Conclusion}\label{sec:conclusions}

In this paper, we apply \emph{fitness landscape analysis} tools
to evaluate the hardness of NAS instances, i.e., the fitness landscape 
of the architecture search optimization problem in a fixed search space, 
and with a defined fitness function and neighborhood operator. 
Given this context, we propose the \emph{fitness landscape footprint}, 
a novel general-purpose framework to characterize and compare NAS problems. 
The insights provided by the footprint may be used to assess the expected performance and to 
relate the difficulties that a search strategy will face
at an instance level, among others.

We evaluate our proposal on instances from two image classification datasets: CIFAR-10 and 
the real-world Remote Sensing dataset of So2Sat LCZ42 for \emph{local climate zone} image classification.
For both, we consider the NAS problem of optimizing convolutional neural networks  
based on the well-known search space of NASBench-101~\cite{ying2019nasbench101}. 

Among the findings, we identify several clues indicating that NAS 
could be performed at shorter regimen (36 epochs), finding elite models early in their training.
Other findings show a common signature of fitness 
of the search space on both datasets and 
the visualization of landscape of both problems for various training settings.

Last but not least, we believe that ability to compare NAS instances using the~\emph{footprint} could help 
(I) identifying the search space generating the most favorable NAS instance out of several possibilities,  
(II)  identifying the fitness evaluation setting (sensor fusion, dataset) generating the most favorable NAS instance out of several possibilities,
(III)  identifying the neighborhood operator (sample encoding, distance or mutation function) generating the most favorable NAS instance out of several possibilities. Or
(IV) identifying a favorable instance using a combination of (I), (II) and or (III).
For future work might investigate some points listed above, as well as the use of insights provided by a~\emph{footprint} 
to help better calibrate a search strategy in a given NAS instance.   



\ifCLASSOPTIONcompsoc
  \section*{Acknowledgments}
        Authors acknowledge support by the European Research Council (ERC) under the European Union's Horizon 2020 research and innovation programme (grant agreement No. [ERC-2016-StG-714087], Acronym: \textit{So2Sat}), by the Helmholtz Association
        through the Framework of Helmholtz AI [grant  number:  ZT-I-PF-5-01] - Local Unit ``Munich Unit @Aeronautics, Space and Transport (MASTr)'' and Helmholtz Excellent Professorship ``Data Science in Earth Observation - Big Data Fusion for Urban Research''(grant number: W2-W3-100),  by the German Federal Ministry of Education and Research (BMBF) in the framework of the international future AI lab "AI4EO -- Artificial Intelligence for Earth Observation: Reasoning, Uncertainties, Ethics and Beyond" (Grant number: 01DD20001) and grant DeToL.
\else
  \section*{Acknowledgment}
\fi

\ifCLASSOPTIONcaptionsoff
  \newpage
\fi

\bibliographystyle{IEEEtran}
\bibliography{bibliography}

\end{document}